\documentclass[10pt,letterpaper]{article}

\usepackage{authblk}
\usepackage{times}
\usepackage{epsfig}
\usepackage{graphicx}
\usepackage[scale=0.85]{geometry}
\usepackage{amsmath}
\usepackage{amssymb}
\usepackage{mathrsfs}
\usepackage{float}
\usepackage{multirow}
\usepackage{xcolor}
\usepackage{algorithm}
\usepackage{algpseudocode}
\usepackage{booktabs}
\usepackage{siunitx}
\usepackage{threeparttable}
\usepackage{multirow}
\usepackage{makecell}      
\usepackage{xcolor}
\usepackage[table]{xcolor}

\usepackage[breaklinks=true,bookmarks=false]{hyperref}
\usepackage{chngcntr}
\counterwithout{table}{section}

\begin{document}

\title{From LIF to QIF: Toward Differentiable Spiking Neurons for Scientific Machine Learning}

\author[1]{Ruyin Wan\thanks{ruyin\_wan@brown.edu}}
\author[2,3]{George Em Karniadakis\thanks{george\_karniadakis@brown.edu}}
\author[2,3,4]{Panos Stinis\thanks{panagiotis.stinis@pnnl.gov}}

\affil[1]{School of Engineering, Brown University, Providence, RI, 02912}
\affil[2]{Pacific Northwest National Laboratory, Richland, WA 99354}
\affil[3]{Division of Applied Mathematics, Brown University, Providence, RI, 02912}
\affil[4]{Department of Applied Mathematics,  University of Washington, Seattle, WA, 98105}

\date{}

\maketitle


\begin{abstract}
Spiking neural networks (SNNs) offer biologically inspired computation but remain underexplored for continuous regression tasks in scientific machine learning. In this work, we introduce and systematically evaluate Quadratic Integrate-and-Fire (QIF) neurons as an alternative to the conventional Leaky Integrate-and-Fire (LIF) model in both directly trained SNNs and ANN-to-SNN conversion frameworks. The QIF neuron exhibits smooth and differentiable spiking dynamics, enabling gradient-based training and stable optimization within architectures such as multilayer perceptrons (MLPs), Deep Operator Networks (DeepONets), and Physics-Informed Neural Networks (PINNs). Across benchmarks on function approximation, operator learning, and partial differential equation (PDE) solving, QIF-based networks yield smoother, more accurate, and more stable predictions than their LIF counterparts, which suffer from discontinuous time-step responses and jagged activation surfaces. These results position the QIF neuron as a computational bridge between spiking and continuous-valued deep learning, advancing the integration of neuroscience-inspired dynamics into physics-informed and operator-learning frameworks.
\end{abstract}

\section{Introduction}

Artificial intelligence (AI) has profoundly influenced the scientific community, emerging as a transformative tool for solving complex problems across diverse domains, including biology~\cite{Dembrower2023ScreeningAI,Jumper2021AlphaFold}, materials science~\cite{Merchant2023Scaling}, and geoscience~\cite{Toms2020GeoscienceNN}. 
Within scientific computing, neural networks—spanning a variety of architectures such as physics-informed neural networks (PINNs)~\cite{raissi2019physics}, Fourier neural operators (FNOs)~\cite{li2021fourier}, and Deep Operator Networks (DeepONets)~\cite{lu2021learning}—have emerged as powerful data-driven solvers and surrogates for physical systems~\cite{jin2020sympnets,zhang2023artificial,liu2024kan}. 
These architectures enable accurate function approximation, solution of partial differential equations (PDEs), and discovery of hidden governing laws that are otherwise intractable with traditional numerical methods.

Despite their success, neural networks remain computationally and energetically expensive, especially when applied to large-scale or highly nonlinear PDE systems~\cite{Lye2019DeepObservables,Zhou2024Unisolver,Tripp2024MeasuringEnergy,Mavromatis2024ComputingWithinLimits,Strubell2021Carbon,Ardakani2016SparseNN}. 
This limitation has motivated growing interest in \textit{spiking neural networks} (SNNs)~\cite{Yan2024Reconsidering,Lemaire2022AnalyticalEnergySNN,Kundu2021SpikeThrift,Wicaksana2021SpikeDyn}, which process information via sparse, event-driven spikes rather than dense, continuous activations. 
SNNs promise reduced energy consumption and enhanced scalability~\cite{maass1997networks,massa2020efficient,kim2022rate}, while being naturally compatible with neuromorphic hardware~\cite{bouvier2019spiking,roy2019towards,davies2018loihi,orchard2021efficient}.

However, the discrete and non-differentiable nature of spike generation presents a significant challenge for gradient-based optimization. 
Conventional approaches such as ANN-to-SNN conversion~\cite{han2020rmp,sengupta2019going,li2021free} attempt to circumvent this by transferring pretrained weights from continuous-valued ANNs to spiking models. 
Although these conversions yield acceptable inference accuracy~\cite{kim2020spiking}, they sacrifice the advantages of true event-driven learning and remain difficult to implement efficiently on neuromorphic devices.
Alternative strategies employ surrogate gradients~\cite{bellec2020solution,neftci2019surrogate} to approximate the derivative of the spike function, enabling direct training, but at the cost of gradient mismatch and unreliable convergence behavior. 
Recently, forward-gradient and local learning methods~\cite{Wan2024RandomizedForwardMode} have sought to replace backpropagation entirely, yet these methods still underperform in accuracy for continuous regression tasks.

Unsupervised schemes such as time-to-first-spike coding, latency-encoding classifiers~\cite{Mostafa2017Temporal}, and spike-timing-dependent plasticity (STDP)~\cite{Bi1998STDP,diehl2015unsupervised} offer biologically plausible alternatives, but they lack a straightforward formulation for regression or PDE-solving tasks that dominate the scientific machine learning (SciML) landscape.

A promising development,  recently proposed by Klos and Memmesheimer~\cite{Klos2025SmoothExactGD}, is the \textit{Quadratic Integrate-and-Fire} (QIF) neuron which is a smooth, differentiable alternative to the conventional Leaky Integrate-and-Fire (LIF) model. 
Unlike the LIF neuron, which produces discontinuous outputs, the QIF model admits a closed-form, continuously differentiable membrane potential near the spiking threshold, enabling exact gradient computation. 
This property makes the QIF neuron uniquely compatible with backpropagation, allowing SNNs to train in a fully differentiable manner while maintaining their event-driven structure.

Such differentiability is particularly valuable for regression and scientific computing, where the fidelity of gradients directly affects the accuracy and stability of solutions to PDEs. 
Traditional LIF-based SNNs often yield jagged, non-smooth predictions that reproduce only the coarse structure of target functions, leading to error accumulation and physically inconsistent dynamics when embedded in multi-layer or physics-informed architectures. 
By contrast, QIF neurons support stable, smooth gradient propagation—an essential prerequisite for scientific regression and operator learning.

In this work, we formalize and extend the application of QIF neurons within spiking neural networks for regression tasks central to SciML. 
Specifically, we integrate QIF dynamics into multiple neural architectures, including multilayer perceptrons (MLPs), DeepONets, and PINNs, and benchmark their performance against LIF-based networks in both direct-training and ANN-to-SNN conversion regimes. 
Through systematic experiments on function approximation, operator learning, and PDE solving, we demonstrate that QIF neurons yield smoother and more accurate predictions, bridging the gap between biologically inspired spiking computation and physics-informed deep learning.

The remainder of this paper is organized as follows:
In Section 2 we review related works on SNN architectures relevant to regression and QIF-based learning.
In Section 3 we formulate the QIF and LIF neuron models and describe their integration within SNN frameworks for regression.
In Section 4 we present experiments on a range of regression and PDE benchmarks, with comparisons against state-of-the-art SNN architectures discussed in Section 2.
Finally, in Section 5 we provide a summary and future research directions.


\section{Related Works}
In this section, we review the most relevant studies on spiking neural networks (SNNs), regression-oriented neural architectures, and their recent integration within SciML. 
The development of spiking multilayer perceptrons (MLPs), spiking deep operator networks (DeepONets) and spiking physics-informed neural networks (PINNs) represents three complementary directions toward extending conventional neural architectures into the spiking domain. 
The spiking MLP serves as a fundamental framework for investigating neural dynamics;
the spiking DeepONet focuses on learning functional mappings between infinite-dimensional spaces, enabling operator learning with spiking dynamics; and the spiking PINN introduces physical constraints into spike-driven models for solving PDEs. 
Together, these models establish a unified foundation study in spike-based learning across both data-driven and physics-informed scientific applications.

\subsection{Spiking neural networks (SNNs)}
Spiking neural networks (SNNs) represent a class of neural architectures that process information through discrete spike events rather than continuous activations~\cite{roy2019towards, Malcolm2023ComprehensiveSNN, Paul2024SurveySNN}. 
Inspired by the signaling mechanisms of biological neurons, SNNs simulate synaptic transmission and membrane potential dynamics to regulate spike generation and information propagation. 
This biologically inspired framework renders SNNs to be more alike neural computation in the brain, while also enabling efficient deployment on neuromorphic hardware platforms \cite{Rathi2023ExploringNeuromorphicComputing, Pritchard2023NeuromorphicReview}. 
Furthermore, the sparse nature of spike-based communication leads to significant advantages in terms of energy efficiency compared to conventional artificial neural networks (ANNs) \cite{Lemaire2022AnalyticalEnergySNN, Wu2022EnergyEfficientSNNFromCNN}.

As illustrated in the bottom-left panel of Fig.~\ref{fig:mlp_pinn_deeponet}, a spiking multilayer perceptron (MLP) operates on streams of input spikes that are processed by layers of spiking neurons to produce output spikes. 
Each neuron integrates incoming spikes by accumulating its membrane potential; once the potential exceeds some potential, the neuron emits a spike and resets its membrane potential. The top-left panel of Fig.~\ref{fig:mlp_pinn_deeponet} depicts two representative spiking neuron dynamics: \textbf{leaky integrate-and-fire (LIF)} and \textbf{quadratic integrate-and-fire (QIF)} neurons. These neurons will be discussed in detail in Section 3.

\subsection{Deep Operator Networks (DeepONets)}

Deep Operator Networks (DeepONets), first proposed by Lu \textit{et al.}~\cite{lu2021learning}, extend neural networks to the learning of \textit{operators}, which are used for mappings between function spaces, rather than approximating a single deterministic function. The key idea is to represent a nonlinear operator through a data-driven combination of two subnetworks: a \textit{branch network} and a \textit{trunk network}. 

The branch network encodes the input function \( f \) by processing its discretized samples or feature representation, while the trunk network takes as input the spatial or temporal coordinate \( y \) where the output function is evaluated. The final output is obtained by coupling the two network outputs through an inner product, effectively expressing the target function as a linear combination of learned basis functions. The overall architecture of the DeepONet framework is illustrated by the blue box in the top-right corner of Fig.~\ref{fig:mlp_pinn_deeponet}.

Mathematically, the approximate operator \( \hat{G} \) can be written as
\[
\hat{\mathcal{G}}(f)(y) = \sum_{i=1}^{p} B_i(f)\, T_i(y),
\]
where \( B_i(f) \) are the coefficients determined by the branch network, and \( T_i(y) \) are the corresponding basis terms provided by the trunk network. Through this formulation, the model learns to approximate a mapping \( \mathcal{G}: \mathcal{F} \rightarrow \mathcal{U} \) between infinite-dimensional spaces, where \( f \in \mathcal{F} \) represents the input function and \( u = \mathcal{G}(f) \in \mathcal{U} \) is the corresponding output function. 

For example, in the context of PDEs, the operator \( \mathcal{G} \) may represent the \textit{solution map} that associates a forcing term \( f \) with the corresponding solution \( u \). By directly learning this operator, DeepONets provide an efficient and generalizable framework for modeling complex physical systems and parameterized PDEs.

\subsubsection{Spiking Deep Operator Networks (Spiking DeepONets)}

A spiking DeepONet can be constructed by replacing either the branch or trunk network of a conventional DeepONet with a SNN~\cite{kahana2022spiking}. 
In this formulation, the original DeepONet is first trained offline as an ANN, and the trained weights are then transferred to the spiking counterpart, where LIF neurons are used in the spiking branch or trunk. 
However, the spiking component in such architectures is typically only partial, since several fully connected layers remain after the spiking stage to generate the final DeepONet outputs~\cite{kahana2022spiking}.

Alternatively, direct training methods based on surrogate gradients can be employed to train spiking DeepONets from scratch. 
In~\cite{Wan2024RandomizedForwardMode}, a spiking DeepONet incorporating LIF neurons demonstrated strong performance using a surrogate-gradient combined with backpropagation-based training schemes. 
In the present work, we compare the performance of QIF neurons against the spiking DeepONet implementation with LIF neuron from~\cite{Wan2024RandomizedForwardMode}. 
The light orange box in the top-right corner of Fig.~\ref{fig:mlp_pinn_deeponet} illustrates the overall structure of the spiking DeepONet, including the spike-based information flow through spiking networks.

\subsection{Physics informed neural networks (PINNs)}

\begin{figure}[H]
    \centering
    \includegraphics[width=0.9\linewidth]{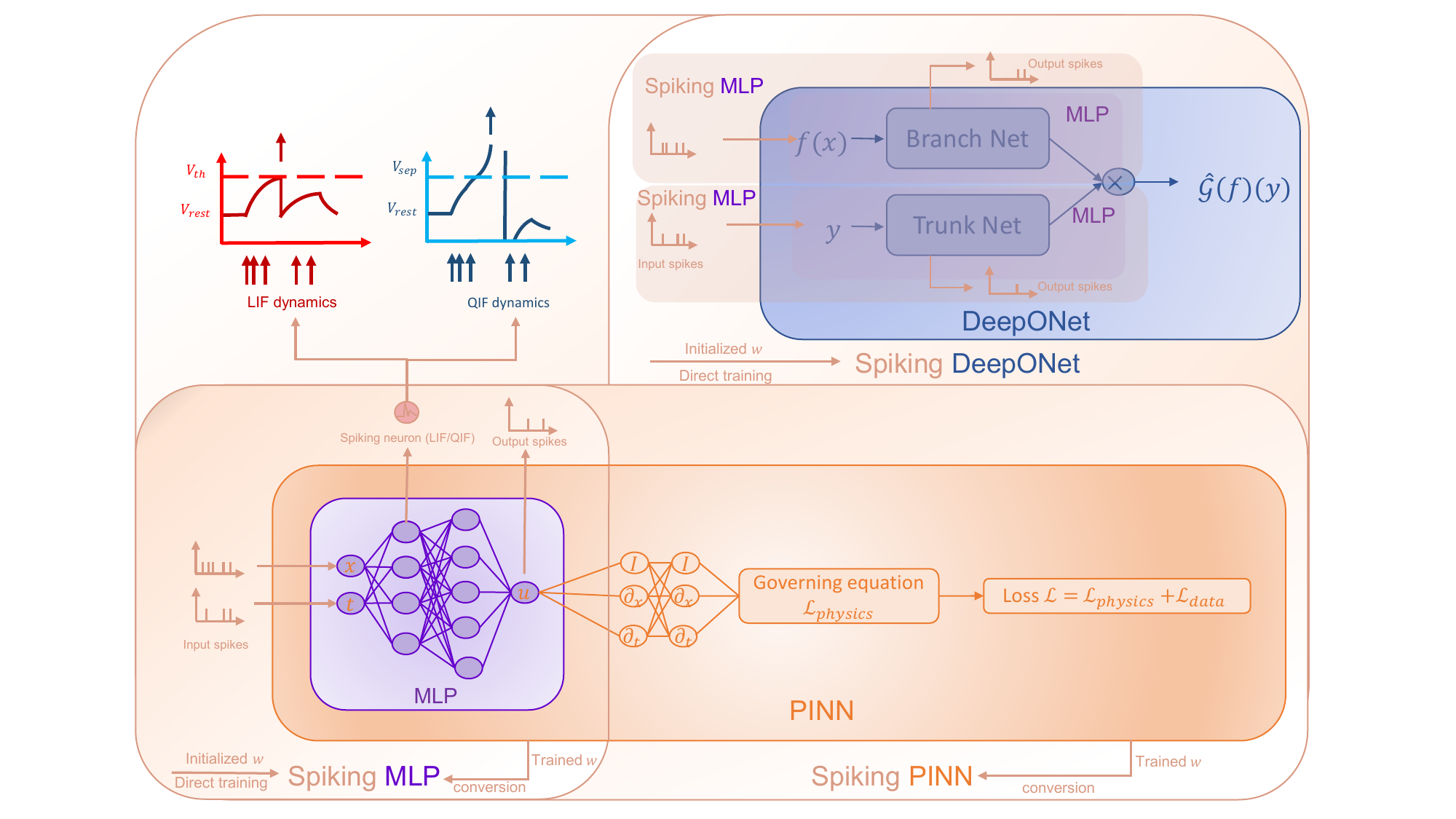}
    \caption{Architectures of MLP, PINN, DeepONet and their corresponding spiking counterparts.}
    \label{fig:mlp_pinn_deeponet}
\end{figure}

Physics-informed neural networks (PINNs) are a class of neural networks that integrate physical laws, expressed as PDEs, directly into the learning process, rather than relying solely on data \cite{raissi2019physics}. This enables PINNs to find solutions that inherently respect the governing physics of a system. By embedding the PDEs into the loss function, PINNs can effectively handle nonlinear and high-dimensional problems, while requiring significantly less training data compared to purely data-driven models \cite{karniadakis2021physics, cuomo2022scientific}. The overall architecture of the PINN framework is illustrated in the bottom panel of Fig.~\ref{fig:mlp_pinn_deeponet}.

The total loss function in a PINN typically consists of two components:
$$\mathcal{L} = \mathcal{L}_\text{physics} + \mathcal{L}_\text{data},$$
where $\mathcal{L}_{\text{physics}}$ enforces the physical constraints by minimizing the PDE residuals at collocation points, and $\mathcal{L}_{\text{data}}$ ensures agreement with available data, such as initial and boundary conditions. In some cases, the initial and boundary conditions can be incorporated analytically (so-called “hard constraints”), allowing the network to focus solely on minimizing the physics loss \cite{liu2022unified}.

By blending numerical analysis with deep learning, PINNs have become a powerful framework for solving forward and inverse problems across various fields in science and engineering \cite{zhang2022analyses, chen2020pinn_nanooptics}. Their ability to generalize beyond limited data makes them particularly well-suited for SciML applications.

\subsubsection{Spiking physics-informed neural networks (Spiking PINNs)}

To extend the capability of PINNs to spiking architectures, a common approach is to first train the PINN as a conventional ANN and subsequently transfer the learned weights to a SNN through ANN-to-SNN conversion~\cite{han2020rmp, sengupta2019going, li2021free, zhang2023artificial}. 
With the incorporation of calibration techniques introduced in~\cite{zhang2023artificial}, the converted spiking PINN can achieve a high level of numerical accuracy that closely matches its ANN counterpart.

However, applying direct training methods to spiking PINNs remains challenging. 
The use of surrogate gradients for spike-based backpropagation often leads to inaccurate gradient estimation, which causes error accumulation during training and multiple gradient computations. For this reason, direct spiking PINNs are not included in our comparative analysis, as a reliable and accurate implementation is currently not feasible.

\section{Methods}

In this section, we first present the governing dynamics of the QIF and LIF neurons, followed by an intuitive discussion highlighting the advantages of QIF over LIF. Finally, we describe the specific adaptations required to implement QIF neurons for regression tasks.

\subsection{QIF and LIF neurons}

The membrane dynamics of the quadratic integrate-and-fire (QIF) neuron are governed by 
\begin{equation}
\dot{V}(t) = V(t)\big(V(t) - 1\big) + I(t),
\label{eq:qif_v}
\end{equation}
\begin{equation}
\tau_s \dot{I}(t) = -\big(I(t) - I_0\big) + \tau_s \sum_i w_i \sum_{t_i} \delta(t - t_i),
\label{eq:i}
\end{equation}
where \( V(t) \) denotes the membrane potential, \( I(t) \) is the input current, \( t_i \) is input spike time at neuron $i$, \( \tau_s \) is the synaptic time constant, and \( w_i \) are the synaptic weights \cite{Klos2025SmoothExactGD}.  

In contrast to Eq.~\eqref{eq:qif_v}, which describes the nonlinear QIF dynamics, the leaky integrate-and-fire (LIF) neuron follows a linear first-order equation with a leakage term:
\begin{equation}
\tau_m \dot{V}(t) = -\lambda \big(V(t) - V_{\text{rest}}\big) + I(t),
\label{eq:lif_v}
\end{equation}
where \( \tau_m \) is the membrane time constant and \( \lambda \) controls the leakage rate of the potential toward its resting value \( V_{\text{rest}} \). 
The current dynamics described by Eq.~\eqref{eq:i} remains identical for both neuron types.

For a LIF neuron, a spike is emitted once the membrane potential exceeds a predefined threshold \( V_{\text{th}} \) and the potential is reset to the resting state \( V_{\text{rest}} \) after the spike. 
The dynamics are plotted by the red trajectory in the top-left panel of Fig.~\ref{fig:mlp_pinn_deeponet}, where all changes remain finite.  

In contrast, the QIF neuron exhibits nonlinear membrane dynamics with a quadratic dependence on \( V(t) \). 
For \( I(t) = 0 \), the system has two equilibrium points: a stable resting potential at \( V_{\text{rest}} = 0 \) and an unstable separatrix at \( V_{\text{sep}} = 1 \). The potential goes toward \( V_{\text{rest}} \) either when  \( V(t) < V_{\text{rest}} \) or \(V_{\text{rest}} < V(t) < V_{\text{sep}} \). Once \( V(t) \) surpasses \( V_{\text{sep}} \), the derivative \( \dot{V}(t) > 0 \), and the potential diverges toward \( +\infty \), indicating a spike event. 
After a spike, \( V(t) \) is instantaneously reset to \( -\infty \), completing the firing cycle. 
This unbounded rise-and-reset behavior is illustrated by the blue trajectory in the top-left panel of Fig.~\ref{fig:mlp_pinn_deeponet}.

\begin{figure}[H]
    \centering
    \includegraphics[width=0.6\linewidth]{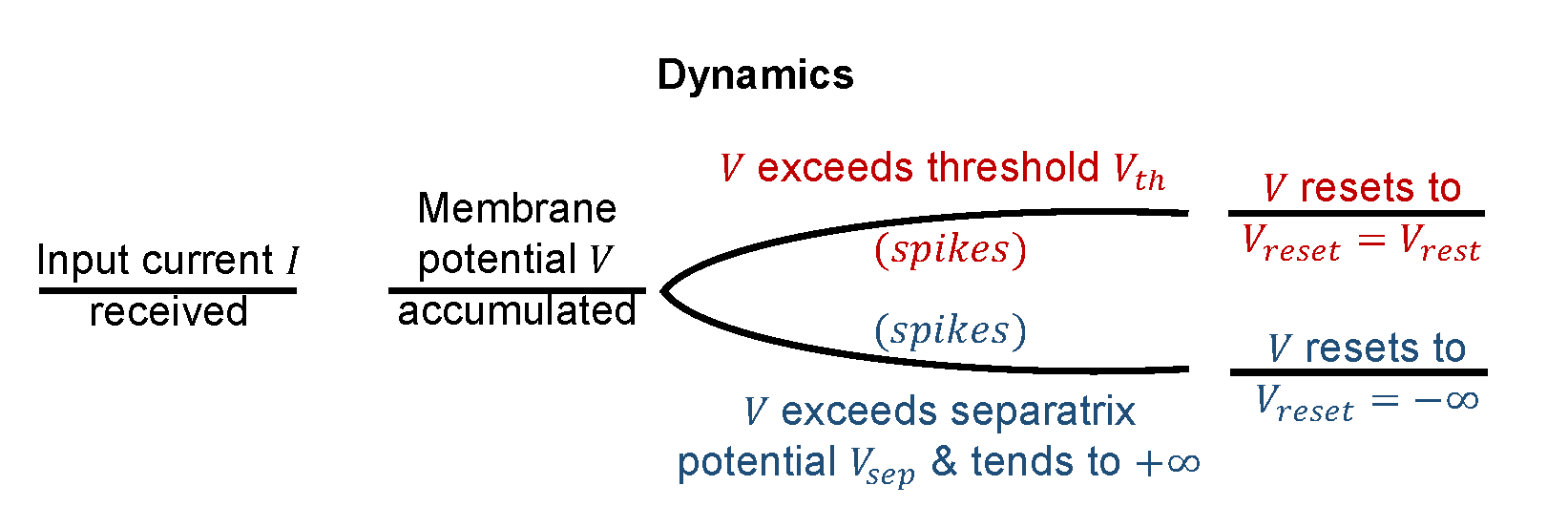}
    \caption{Spike generation and membrane potential dynamics of LIF (red) and QIF (blue) neurons.}
    \label{fig:lif_qif_dynamics}
\end{figure}

Fig.~\ref{fig:lif_qif_dynamics} is a flow chart of LIF and QIF neuron spiking dynamics. Here and in the
following, a trial refers to a finite duration individual simulation of a system of neurons.

\subsection{Advantages of QIF over LIF}

QIF neurons possess inherently smooth dynamics that allow the variation of membrane potential \( \dot{V}(t) \) to grow unbounded during spike generation. 
As a result, small perturbations in synaptic weights \( w \) or input spike times \( t \) influence only the precise timing of a spike rather than determining whether a spike occurs at all. If any change has shifted an output spike time outside the trial, i.e. an output spike disappears at the trial end, this spike becomes a \textit{pseudospike} and \textit{pseudodynamics} is implemented to model this (see Section 3.3.4). This introduction of \textit{pseudospikes} and \textit{pseudodynamics} avoids the model's expressivity limitation of not generating enough spikes within the trial~\cite{Klos2025SmoothExactGD}. This continuous and non-disruptive spike timing behavior yields a smooth gradient of output spikes with respect to both \( w \) and \( t \), which is essential for gradient-based optimization and learning (see Sections 3.3.2 and 3.3.3). 

In contrast, the LIF neuron exhibits discontinuous spike generation dynamics. 
This is because minor variations in \( w \) or \( t \) can change whether a spike occurs, leading to discontinuities in the spike train and making the gradient undefined at spike events. 
Consequently, exact gradient computation is not possible for LIF neurons, and surrogate or approximate gradient methods are required for training.

\subsection{Oscillatory QIF neurons with infinitesimally short coupling}

In the present work, we focus on the implementation and adaptation of the QIF neuron for regression tasks and PDE learning. While the QIF neuron with extended coupling (Eqs. \ref{eq:qif_v} and \ref{eq:i}) might give a better result for regression problems, we use oscillatory QIF neurons with infinitesimally short coupling for simplicity and better fitting for an event-driven framework.
In this section, we give a simple formulation and intuition towards the smoothness of the oscillatory QIF neuron which used in our examples. In all following proofs, we constrain to the simplest case where event order is fixed and no event coincidence occurs. For the detailed mathematical formulation of the QIF neuron with extended coupling, its dynamic properties, and the theoretical proof of gradient smoothness, including scenarios where event order changes or event coincidence happens, please see~\cite{Klos2025SmoothExactGD}.  

\subsubsection{Ordinary spike dynamics}
For oscillatory QIF neurons, the membrane potential \(V(t)\) evolves according to
\begin{equation}
\tau_m \dot{V}(t) = V(t)\big(V(t) - 1\big) + I(t),
\label{eq:oqif_v}
\end{equation}
where the input current \(I(t)\) consists of a constant drive \(I_0\) and instantaneous delta-coupled synaptic inputs,
\begin{equation}
I(t) = I_0 + \tau_m \sum_i w_i \sum_{t_i} \delta(t - t_i).
\label{eq:oqif_i}
\end{equation}
Here, each incoming pre-synaptic spike at time \(t_i\) causes an instantaneous jump of the membrane potential \(V\) by the synaptic weight \(w_i\), while the subthreshold dynamics between spikes follow Eq.~\eqref{eq:oqif_v}. For a chosen suprathreshold constant current \(I_0\), the membrane potential increases monotonically, leading to self-sustained oscillations even in the absence of external input.

To describe these oscillations uniformly, the voltage dynamics is mapped to a phase representation using
\begin{equation}
\phi = \Phi(V) =
\frac{\tau_m}{\sqrt{I_0 - \tfrac{1}{4}}}
\left[
\arctan\!\left(\frac{V - \tfrac{1}{2}}{\sqrt{I_0 - \tfrac{1}{4}}}\right)
+ \frac{\pi}{2}
\right],
\label{eq:oqif_phase}
\end{equation}
which transforms the QIF neuron into an equivalent $\Theta$-neuron \cite{Mirollo1990, Viriyopase2018}. In this representation, the phase evolves with constant velocity between spike arrivals,
\begin{equation}
\dot{\phi}(t) = 1,
\label{eq:oqif_dphi}
\end{equation}
and the threshold and reset correspond to
\begin{equation}
\phi_\Theta = \Phi(\infty) = \frac{\tau_m \pi}{\sqrt{I_0 - \tfrac{1}{4}}},
\qquad
\phi_\mathrm{reset} = \Phi(-\infty) = 0.
\label{eq:oqif_bounds}
\end{equation}
If the initial phase is \(\phi(0) = \phi_0\), the free evolution of the neuron is simply
\begin{equation}
\phi(t) = \phi_0 + t.
\label{eq:oqif_solution}
\end{equation}

\begin{figure}[H]
    \centering
    \includegraphics[width=0.4\linewidth]{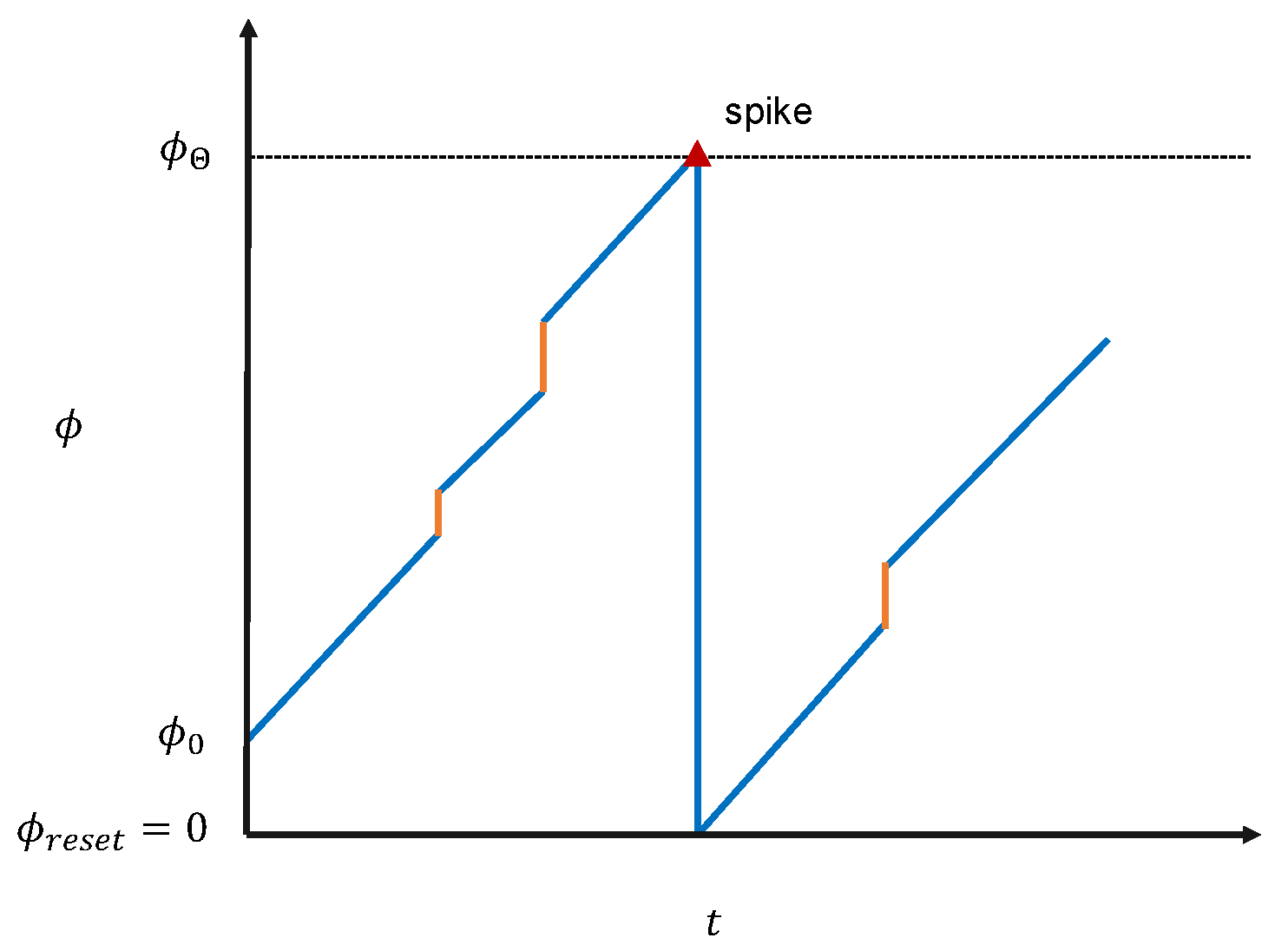}
    \caption{Plot of phase $\phi$ with respect to time $t$. $\phi(t)$ is in $[0,\phi_\Theta)$. The phase changes linearly between spike arrivals and experiences an instantaneous jump when a spike arrives. Once the threshold $\phi_\Theta$ is reached, a spike is emitted and phase reset to circle 'starting' point at $\phi_\text{reset}=0.$}
    \label{fig:phase_plot}
\end{figure}

Upon the arrival of an input spike at time \(t_i\), the phase undergoes an instantaneous transition governed by the \emph{phase transition curve} (PTC),
\begin{equation}
\phi(t_i^+) = H_{w_i}\!\big(\phi(t_i^-)\big)
= \Phi\!\left(\Phi^{-1}\!\big(\phi(t_i^-)\big) + w_i\right),
\label{eq:oqif_ptc}
\end{equation}
which is smooth with respect to the synaptic weight \(w_i\) as well as input spike time $t_i$. Between events, the phase increases linearly as in Eq.~\eqref{eq:oqif_dphi}, while each spike arrival instantaneously shifts its trajectory on the unit circle according to \(H_{w_i}\). Fig. \ref{fig:phase_plot} illustrate the phase change, in which a spike is generated when pass a finite $\phi_\Theta$ and reset to $\phi_\text{reset}=0$, avoiding infinities.

\subsubsection{Smoothness of spike time: single input}
For a single pre-synaptic spike from neuron $i$ at time \(t_i\), the postsynaptic neuron reaches threshold \(\phi_\Theta\) at time \(t_{\mathrm{sp}}\), satisfying
\begin{equation}
\phi_\Theta = H_{w_i}\!\big(\phi(t_i^-)\big) + (t_{\mathrm{sp}} - t_i).
\end{equation}
Thus, the output spike time depends on the synaptic weight as
\begin{equation}
t_{\mathrm{sp}}(w_i, t_i) = t_i + \phi_\Theta - H_{w_i}\!\big(\phi(t_i^-)\big),
\label{eq:tsp_single}
\end{equation}
and its derivatives with respect to weights and spike input times given by 
\begin{equation}
\frac{\partial t_{\mathrm{sp}}}{\partial w_i}
= -\frac{\partial H_{w_i}\!\big(\phi_0 + t_i\big)}{\partial w_i} \;\; \text{and} \;\;
\frac{\partial t_{\mathrm{sp}}}{\partial t_i}
= 1-\frac{\partial H_{w_i}\!\big(\phi_0+t_i\big)}{\partial t_i},
\label{eq:tsp_grad}
\end{equation}
are continuous due to the smoothness of \(H_{w_i}\).

\subsubsection{Smoothness of spike time: multiple inputs}
For multiple pre-synaptic spikes at times
\(t_{i_1} < t_{i_2} < \dots < t_{i_m} < t_{\mathrm{sp}}\),
the composite phase after the last input is given by
\begin{equation}
\phi(t_{i_m}^+) =
H_{w_{i_m}} \circ H_{w_{i_{m-1}}} \circ \dots \circ H_{w_{i_1}}\!\big(\phi(t_{i_1}^-)\big),
\end{equation}
and the corresponding output spike time becomes
\begin{equation}
t_{\mathrm{sp}}(\mathbf{w}, \mathbf{t}) =
t_{i_m} + \phi_\Theta -
H_{w_{i_m}} \circ H_{w_{i_{m-1}}} \circ \dots \circ H_{w_{i_1}}\!\big(\phi(t_{i_1}^-)\big)
\end{equation}
where $\mathbf{w} = (w_1, w_2, \cdots, w_N)$ corresponds to all $N$ presynaptic neurons connecting to the postsynaptic neuron and $\mathbf{t}=\{t_{i_{k}}| i =1,..,N; \text{$k$ indexes spikes from neuron $i$} \}$. During one postsynaptic interspike interval, only $m$ of these events actually arrive before the next postsynaptic spike at time $t_{sp}$, so the subscript $i_k$ refers to which presynaptic neuron fired the $k^{th}$ spike.

By the chain rule, the sensitivity of the spike time with respect to each weight is
\begin{equation}
\frac{\partial t_{\mathrm{sp}}}{\partial w_{i_k}}
= -\!\left(\prod_{\ell > k} H'_{w_{i_\ell}}(\cdot)\right)
\frac{\partial H_{w_{i_k}}(\cdot)}{\partial w_{i_k}}.
\end{equation}
For differentiation with respect to input spike time, we define the recursive relations
\begin{align}
\xi_1 &= \phi_0 + t_{i_1}, 
&\psi_1 &= H_{w_{i_1}}(\xi_1), \\
\xi_k &= \psi_{k-1} + (t_{i_k} - t_{i_{k-1}}), \quad k \ge 2, 
&\psi_k &= H_{w_{i_k}}(\xi_k),
\end{align}
such that the phase immediately after the last input spike is $\phi(t_{i_m}^+) = \psi_m$.
The corresponding output spike time is then
\begin{equation}
t_{\mathrm{sp}}(\mathbf{t}, \mathbf{w})
= t_{i_m} + \phi_\Theta - \psi_m.
\label{eq:tsp_multi_inputs}
\end{equation}

Let $A_k = H'_{w_{i_k}}(\xi_k)$ and define the partial products $P_{a:b} = \prod_{\ell=a}^{b} A_\ell$ (with empty product $P_{a:b}=1$ if $a>b$). 
Applying the chain rule yields the derivative of $\psi_m$ with respect to each input time $t_{i_k}$:
\begin{equation}
\frac{\partial \psi_m}{\partial t_{i_k}}
= P_{k+1:m}\,A_k - P_{k+2:m}\,A_{k+1},
\label{eq:dpsim_dtik}
\end{equation}
and consequently,
\begin{equation}
\frac{\partial t_{\mathrm{sp}}}{\partial t_{i_k}}
= \delta_{k,m} - \frac{\partial \psi_m}{\partial t_{i_k}}
= \delta_{k,m} - \Big(P_{k+1:m}\,A_k - P_{k+2:m}\,A_{k+1}\Big),
\label{eq:dtsp_dtik}
\end{equation}
where $\delta_{k,m}$ is the Kronecker delta.

All quantities in Eqs.~\eqref{eq:dpsim_dtik}--\eqref{eq:dtsp_dtik} are smooth functions of input times $\mathbf{t}$,
as long as the event order 
$t_{i_1} < \cdots < t_{i_m} < t_{\mathrm{sp}}$
remains fixed. 

Hence, despite the jump-like voltage response at each spike arrival, the output spike time remains a piecewise smooth function of the synaptic weights and input times. This property ensures that gradient-based optimization methods remain valid even in spiking networks with delta-coupled synaptic interactions.

\subsubsection{Pseudospike dynamics}
At the end of ordinary trial $T$, where the voltage might not have reached the next threshold crossing, the neuron still contains latent evidence toward firing, encoded in its membrane potential $V(T)$. To make the gradient smooth, we continue to compute the latent trajectory beyond $T$, by defining the pseudovoltage:
\begin{equation}
\tau_m \dot{V}_{ps}(t) = V_{ps}(t)\big(V_{ps}(t) - 1\big) + I_0
\end{equation}
with post-trial initial condition 
\begin{equation}
V_{ps}(T) = V(T)+\sum_j w_j \frac{\Phi(V_{ps,j}(T))}{\phi_\Theta}.
\label{eq:psedo-ini}
\end{equation}
The summation over \( j \) accounts for all presynaptic neurons connected to the postsynaptic neuron. The second term in Eq.~\eqref{eq:psedo-ini} is not a literal voltage contribution, but rather a smooth surrogate for the synaptic input that would occur immediately after the trial end at \(T\). It takes the form of a weighted average of normalized presynaptic phases, where each ratio \(\Phi(V_{ps,j}(T)) / \phi_{\Theta}\) provides a continuous measure of how close each presynaptic neuron is to completing its current cycle, and thereby how strongly it would influence the postsynaptic neuron if the simulation were extended. In this way, the post-trial initial condition avoids an artificial cutoff of coupling effects at the trial boundary.

The $k$-th spike time, if it is a pseudospike, is then given by
\begin{equation}
t_{ps} = T + (k - n_{\mathrm{trial}})\phi_\Theta
- H\!\big(\phi_T, u\big),
\label{eq:pseudo_time}
\end{equation}
where 
\begin{equation}
H\!\big(\phi, u\big) = \Phi\!\left(\Phi^{-1}\!\big(\phi \big) + u\right), 
\qquad
\phi_T = \Phi(V(T)),
\qquad
u = \sum_j w_j \frac{\Phi(V_{ps,j}(T))}{\phi_\Theta}.
\end{equation}
$n_{\mathrm{trial}}$ denotes the number of ordinary spikes before the trial end.

Define the normalized pseudostates
\begin{equation}
r_i = \frac{\Phi(V_{ps,i}(T))}{\phi_\Theta},
\label{eq:pseudo_r}
\end{equation}
which satisfy the one-shot “rate-like” coupling equations
\begin{equation}
r_i = \frac{\Phi\!\left(V_i(T) + \sum_j w_j r_j\right)}{\phi_\Theta}
=: f_i\!\left(\sum_j w_j r_j\right),
\label{eq:pseudo_r_equation}
\end{equation}
and let $u = \sum_j w_j r_j$. This setup makes all pseudospike quantities explicit in terms of the terminal voltages $V_i(T)$ and the weights $\mathbf{w}$. 

With the above definitions, it is easy to show that the pseudospike time $t_{ps}$ is smooth with respect to weight and input time when no network spike crosses the trial end \cite{Klos2025SmoothExactGD}. 

\subsubsection{Smoothness at T: ordinary-pseudo spike transition at trial end}

We now verify that both the pseudospike times and their gradients remain continuous at the trial end \(t=T\), where transitions occur between ordinary and pseudospikes. 

\textbf{(i) Value continuity.}  
When the last ordinary spike ($k^{th}$ spike) time approaches the trial end $T$, i.e. $t_{sp} \nearrow T$, we have $V(T)\to -\infty$ since reset happens within the trial. Hence, 
\begin{equation}
    t_{ps} = T + (k - k)\phi_\Theta- H\!\big(\Phi(-\infty), u\big)= T + \Phi\!\left(\Phi^{-1}\!\big(0\big) + u\right)= T + \Phi\!\left(-\infty\right) = T.
\end{equation}
When the first pseudospike ($k^{th}$ spike) time approaches the trial end $T$, i.e. $t_{ps} \searrow T$, we have $V(T)\to \infty$ since reset does not happen within the trial. Hence, 
\begin{equation}
    t_{sp} = T + \phi_\Theta- H_{w}\!\big(\Phi(\infty)\big) = T.
\end{equation}
Thus, continuity at the transition is ensured.

\textbf{(ii) Gradient continuity.}  
Differentiating Eq.~\eqref{eq:pseudo_time} with respect to any scalar parameter \(p\) (weight or input time) gives
\begin{equation}
\frac{\partial t_{ps}}{\partial p}
=-\big(H_\phi(\phi_T,u)\, \frac{\partial \phi_T}{\partial p}
+H_u(\phi_T,u)\, \frac{\partial u}{\partial p}\big),
\label{eq:dtps_general}
\end{equation}
where
\begin{equation}
H_\phi(\phi,u) = \Phi'(\Phi^{-1}(\phi)+u) \cdot \frac{d}{d\phi} (\Phi^{-1}(\phi))
=\frac{\Phi'(\Phi^{-1}(\phi_T)+u)}{\Phi'(\Phi^{-1}(\phi_T))}
\;\; \text{and} \;\;
H_u(\phi_T,u)=\Phi'(\Phi^{-1}(\phi_T)+u),
\end{equation}
since 
\begin{equation}
\frac{d}{d\phi} [\Phi(\Phi^{-1}(\phi))] = \frac{d}{d\phi} [\phi] = 1
\Rightarrow \Phi'(\Phi^{-1}(\phi)) \cdot \frac{d}{d\phi} [\Phi^{-1}(\phi)] =1
\Rightarrow
\frac{d}{d\phi} (\Phi^{-1}(\phi))
=\frac{1}{\Phi'(\Phi^{-1}(\phi_T))}.
\end{equation}

The last 'ordinary' point where we evaluate the ODE is
\(t_{\mathrm{ord}}=t_{i_k}+\phi_\Theta-\psi_k\),
with post-jump phase \(\psi_k=H(\phi_T-\varepsilon,u)\),
where \(\varepsilon=T-t_{\mathrm{ord}}\!\to\!0\). Note that $t_{ik}$ is the last actual input spike before the boundary T, but $t_{ord}$ is a hypothetical point. In transition, $t_{ik}$ tends to $T$, so we also write $t_{ik}=T-\varepsilon$ for simplicity. Expanding \(H(\phi_T-\varepsilon,u)
=H(\phi_T,u)-H_\phi(\phi_T,u)\,\varepsilon+o(\varepsilon)\)
and substituting it together with $t_{ik} = T - \varepsilon$ yields
\begin{equation}
t_{\mathrm{ord}}
=T+\phi_\Theta-H(\phi_T,u)
+\big(H_\phi(\phi_T,u)-1\big)\varepsilon+o(\varepsilon).
\end{equation}
Differentiating with respect to \(p\),
\begin{equation}
\frac{\partial t_{\mathrm{ord}}}{\partial p}
=-\big(H_\phi(\phi_T,u)\, \frac{\partial \phi_T}{\partial p}
+H_u(\phi_T,u)\, \frac{\partial u}{\partial p}\big) + \big(H_\phi(\phi_T,u)-1\big)\frac{\partial \varepsilon }{\partial p}+ o(1).
\end{equation}
Using
\(\frac{\partial \varepsilon}{\partial p}=-\frac{\partial t_{\mathrm{ord}}}{\partial p}\),
we obtain
\begin{equation}
H_{\phi}\frac{\partial t_{\mathrm{ord}}}{\partial p}
=-\big(H_\phi(\phi_T,u)\, \frac{\partial \phi_T}{\partial p}
+H_u(\phi_T,u)\, \frac{\partial u}{\partial p}\big) + o(1).
\end{equation}
Dividing by $H_{\phi}$,
\begin{equation}
\frac{\partial t_{\mathrm{ord}}}{\partial p}
=-(\frac{\partial \phi_T}{\partial p}
+\frac{H_u}{H_{\phi}}\,\frac{\partial u}{\partial p}) +o(1).
\label{eq:37}
\end{equation}
At the limit $\varepsilon \rightarrow 0$, $V(T)\rightarrow+\infty$, with
\begin{equation}
H_\phi(\phi_T,u) =\frac{\Phi'(V(T)+u)}{\Phi'(V(T))}
\;\; \text{and} \;\;
\Phi'(V) = \frac{\tau_m}{(I_0 - \frac{1}{4})+(V-\frac{1}{2})^2},
\end{equation}
the limit 
\begin{equation}
H_\phi(\phi_T,u) = \frac{{(I_0 - \frac{1}{4})+(V_T-\frac{1}{2})^2}}{{(I_0 - \frac{1}{4})+(V_T+u-\frac{1}{2})^2}} = \frac{\frac{(I_0 - \frac{1}{4})}{V_T^2}+(1-\frac{1}{2V_T})^2}{\frac{(I_0 - \frac{1}{4})}{V_T^2}+(1-\frac{u-\frac{1}{2}}{V_T})^2} \rightarrow 1.
\end{equation}
Therefore, Eq. \ref{eq:37} becomes
\begin{equation}
\frac{\partial t_{\mathrm{ord}}}{\partial p}
=-(H_{\phi}(\phi_T,u)\frac{\partial \phi_T}{\partial p}
+H_u(\phi_T,u)\,\frac{\partial u}{\partial p})
\end{equation}
at the limit and this coincide with that of $t_{ps}$.
Hence, the gradient is continuous through the trial boundary with respect to both
weights and input times.

\paragraph{Note: In our simulations, we follow the same setup as in \cite{Klos2025SmoothExactGD} (see Section I.C in Supplementary Material there) and we ignore infinitesimal effects extremely close to threshold/reset. Concretely, if an incoming spike arrives when $\phi>\phi_\Theta-\varepsilon$ or $\phi<\varepsilon$ with $\varepsilon=10^{-6}$, we neglect its effect on $\phi$ and thus on the next spike time.}

\subsection{QIF neurons for regression tasks}

The QIF neuron model, introduced in \cite{Klos2025SmoothExactGD}, provides a mathematically smooth formulation of spike generation within an event-driven computational framework. Unlike conventional time-step-based SNNs, the event-driven formulation allows the dynamics to evolve in time and the continuous time is the main focus of the dynamics. This is more naturally suited for QIF neuron with its differentiability with respect to time. Since regression tasks in SNNs are traditionally treated under the time-stepped paradigm, it is necessary to establish how continuous inputs and outputs can be consistently represented in the event-driven QIF framework.

\subsubsection{Encoding methods}

In time-stepped SNNs, continuous inputs are typically converted into spike trains using encoding schemes such as rate encoding, temporal encoding~\cite{Guo2021NeuralCodingSNN}, floating-point encoding, or lower-triangular encoding~\cite{zhang2023artificial}. However, encodings that yield discrete 0–1 spike representations (e.g., rate or lower-triangular encodings) are incompatible with the event-driven formulation, where spike timing rather than spike count conveys the information. Moreover, these discrete encodings disrupt differentiability, which is essential in physics-informed neural networks (PINNs) that require derivatives of the network outputs with respect to the inputs.

For classification tasks such as MNIST~\cite{Klos2025SmoothExactGD}, temporal encoding can be used to map pixel intensities to spike times across neuron indices. This approach is effective for classification but unsuitable for regression, where each input point carries equal importance and cannot be prioritized by intensity. An alternative is to employ Gaussian receptive field (GRF) encoding, where an input value \(X\) is represented by the spike times of \(m\) neurons with receptive centers \(c_i\):
\begin{align}
t_i &= \left(1 - \frac{r}{\max(r)}\right)T, \\
r &= \exp\!\left(-\frac{(X - c_i)^2}{2\sigma^2}\right), \\
c_i &= X_{\min} + i\frac{X_{\max} - X_{\min}}{m}, \\
\sigma &= \frac{X_{\max} - X_{\min}}{\beta(m - 2)},
\end{align}
where \(T\) denotes the trial duration, $\sigma$ controls the width of the GRF field and the scalar $\beta >0$ is a scaling coefficient adjusting the amount of overlap between adjacent receptive fields. The derivative of the network output \(u\) with respect to \(x\) can then be expressed as
\begin{equation}
\frac{du}{dX} = \sum_{i=0}^{m-1} \frac{du}{dt_i} \frac{dt_i}{dX}.
\end{equation}
The QIF neuron ensures that \(u\) remains differentiable with respect to spike times \(t_i\)~\cite{Klos2025SmoothExactGD}, while the encoding formulation guarantees that \(\frac{dt_i}{dX}\) is smooth. Despite these advantages, this method becomes computationally expensive for large \(m\), and a small \(m\) leads to a significant degradation in performance. Consequently, we propose using unencoded inputs directly, interpreting them as spike times. To achieve this, inputs are first normalized to the interval \([0,1]\) and then transformed into spike times distributed across some time $t_\text{trans}$ (which could be chosen to be the trial duration \(T\)), matching the idea that larger values spike first:
\begin{equation}
X^{(t)} = \left(1 - \frac{X - X_{\min}}{X_{\max} - X_{\min}}\right)t_\text{trans}.
\end{equation}
This formulation preserves differentiability and requires only a single differentiation per input, resulting in a substantial reduction in computational cost compared to multi-neuron encodings.

\subsubsection{Output spike interpretation}

In event-driven QIF networks, outputs are expressed as spike times, which can make it difficult to represent small or negative target values since the membrane potential must integrate sufficiently before a spike is emitted. This limitation poses challenges in regression tasks that frequently involve small, zero-valued or negative outputs. To mitigate this, we define the network output as the interval between the first output spike time for two output neurons rather than the absolute timing for the first spike of a single output neuron:
\begin{equation}
f(x) = t_{\text{2,out}} - t_{\text{1,out}}.
\end{equation}
This temporal-difference formulation yields a smooth mapping between input and output, while maintaining the event-driven nature of the QIF dynamics and ensuring compatibility with gradient-based optimization.




\section{Experiments}

To evaluate the effectiveness of QIF neurons in regression tasks, we conduct experiments across three representative settings: function regression using MLPs, operator learning with DeepONets. and PDE solving with PINNs. For comparison, we also test models employing leaky integrate-and-fire (LIF) neurons with different simulation step counts. Specifically, \textbf{LIF-32}, \textbf{LIF-64}, and \textbf{LIF-128} denote models using 32, 64, and 128 simulation time steps, respectively. These simulation time steps determine the temporal resolution of membrane potential updates during spike evolution, but do not correspond to the number of training data. In addition to direct ANN-to-SNN conversion, we adopt the calibration method proposed in \cite{zhang2023artificial} to further enhance accuracy. We denote directly converted models by ``\textbf{(conv)}'' and calibrated models by ``\textbf{(cal)}'' throughout the experiments. To compare the various approaches, we use the $L_2$ relative error defined as

\begin{equation}
\epsilon_{L_2} = \frac{\|u - u_E\|_2}{\|u_E\|_2}, 
\end{equation}
where ${\|\cdot \|_2}$ denotes the $L_2$ norm, and $u$ and $u_E$ are the  predicted and reference solutions on the test set, respectively. 

\subsection{Function regression with MLP}

We first evaluate the performance of QIF neurons on function regression tasks through several one-dimensional and two-dimensional examples.

\subsubsection{Parabola}
As a mini test case, we consider the simple parabola $u(x) = x^2$ to illustrate the smoothness and convergence behavior of QIF neurons over training epochs. The network architecture consists of four hidden layers with 64 neurons each ($4\times64$). A total of 100 $(x, u(x))$ samples are used for training and 1000 for testing. The simulation duration for QIF neurons is set to $T = 2.0$, and the models are trained for 10{,}000 epochs to monitor the evolution of loss and training time.

\begin{figure}[H]
\centering
    \begin{tabular}{ccc}
     \includegraphics[width=0.28\textwidth]{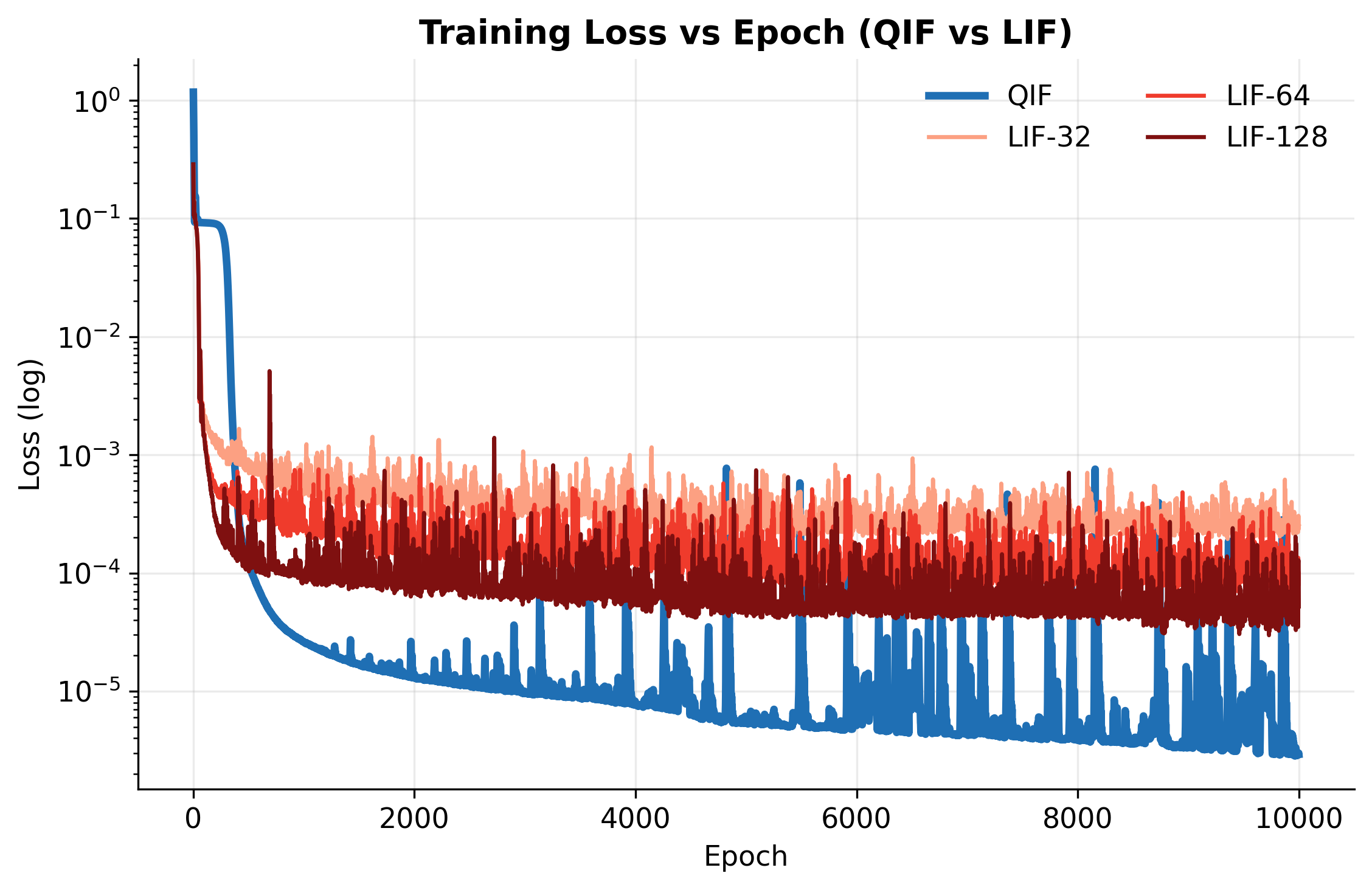} &
     \includegraphics[width=0.35\textwidth]{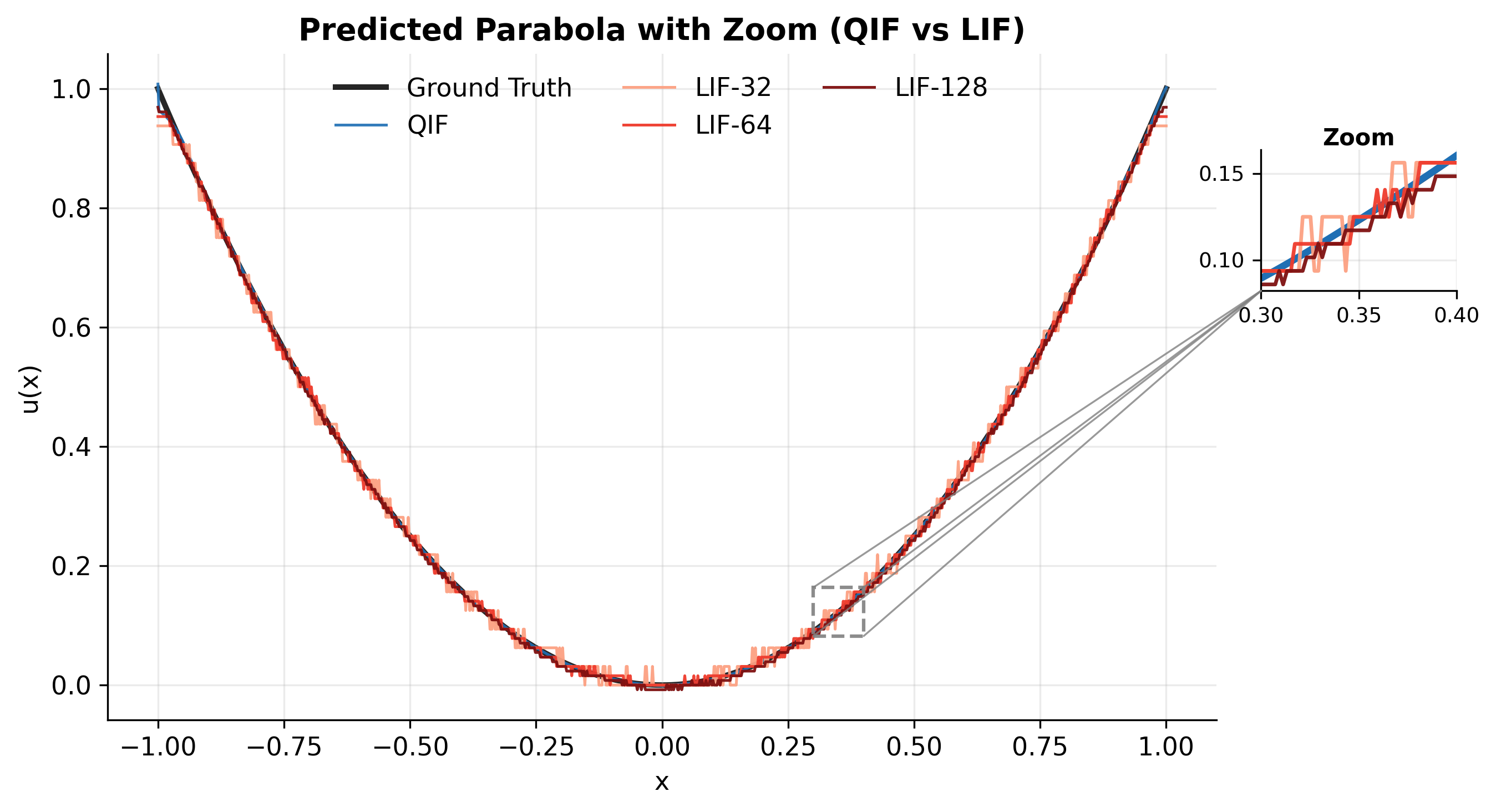} &
\includegraphics[width=0.37\textwidth]{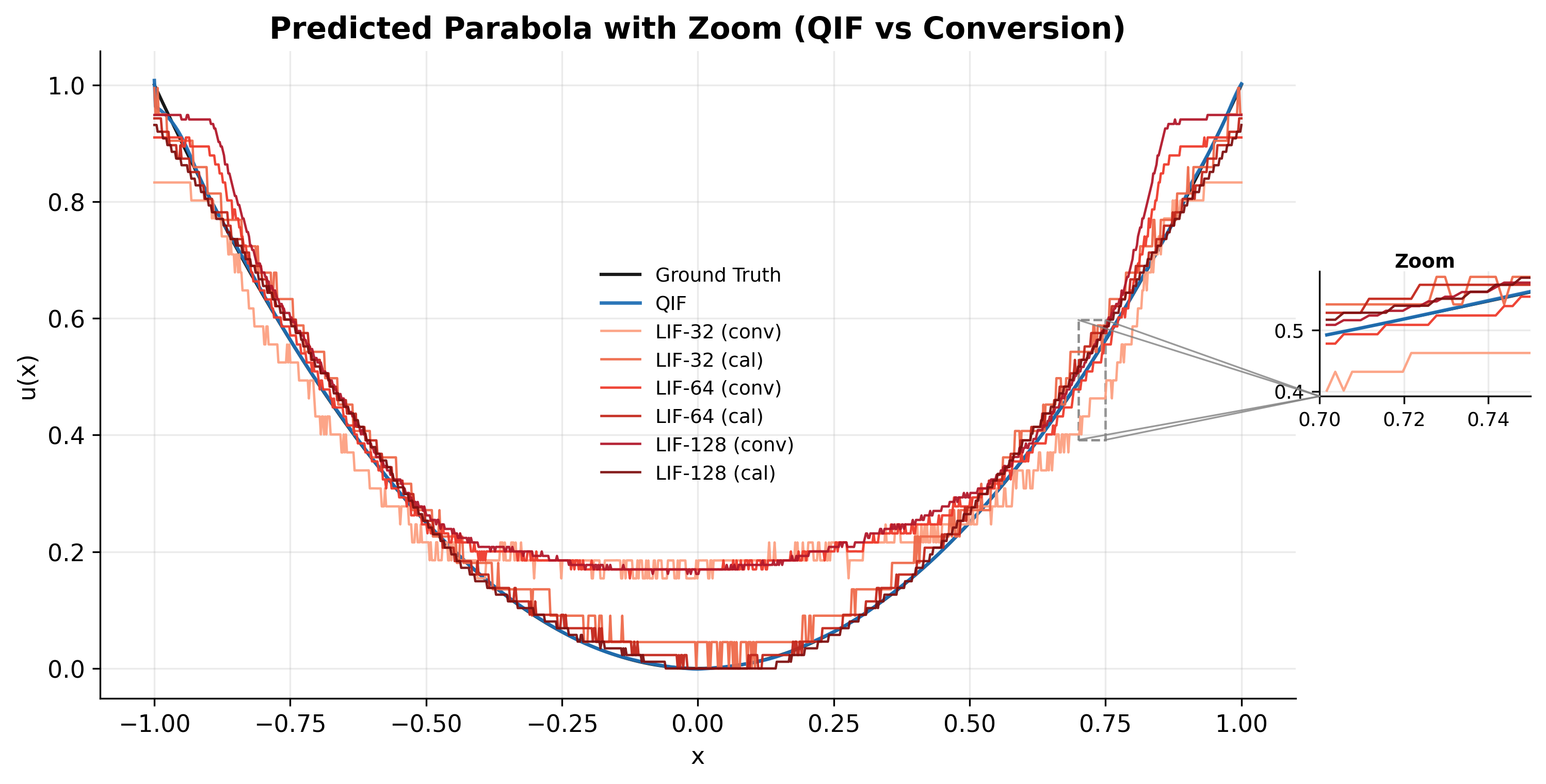} \\
  (a) Training loss history & (b) LIF direct training and QIF & (c) LIF conversion and QIF \\
\end{tabular}
  \caption{Parabola results: (a): Loss history in training by QIF neuron compared with direct SNN training by LIF neuron. (b) \& (c): Parabola ground truth and predictions using QIF and LIF neurons. }
\label{fig:parabola}
\end{figure}

As shown in Fig.~\ref{fig:parabola}(a), for directly trained LIF-based networks, the training loss decreases as the number of simulation steps increases, but this improvement comes at the expense of significantly higher computational cost. In contrast, the QIF neuron achieves the lowest loss with a much smoother convergence profile during training. For comparison, ANN-to-SNN conversion is also performed using the same network setup. As illustrated in Fig.~\ref{fig:parabola}(c), the converted models without calibration even fail to accurately reproduce the parabola, giving relative $L_2$ error of more than 20\%, so the prediction quality does not improve with an increased number of simulation steps.

From Table~\ref{tab:metrics}, the QIF model (highlighted in light blue) achieves the lowest error across all tested configurations. The next best performance is obtained by the directly trained SNN using the LIF neuron with 128 simulation steps, but even in this case, its relative $L_2$ error exceeds that of the QIF model by more than a factor of two. For converted models that fail to produce accurate predictions, the error increases dramatically, which are up to approximately 36 times larger than the QIF error. Even with calibration of the converted LIF model, the best error is still more than $7\times$ the QIF error. 

Figures~\ref{fig:parabola}(b) and (c) further compare the output smoothness of directly trained QIF and LIF models, as well as the converted SNNs. Although both neuron types roughly capture the parabola shape, the LIF-based predictions exhibit noticeable jaggedness, which is clearly visible in the zoomed-in plots, even when using 128 simulation steps and applying calibration after conversion. In contrast, the QIF neuron produces smooth and stable outputs, demonstrating its advantage in regression tasks requiring continuity and differentiability.

\subsubsection{Ricker wavelet}

We further evaluate the QIF neuron on a two-dimensional function fitting problem using the Ricker wavelet, defined as
\[
u(x, y) = \frac{1}{\pi \sigma^4} \left(1 - \frac{1}{2}\frac{x^2 + y^2}{\sigma^2}\right) 
\exp\left(-\frac{1}{2}\frac{x^2 + y^2}{\sigma^2}\right),
\]
where $\sigma = 0.8$. This function serves as a smooth, radially symmetric benchmark for testing spatial regression capabilities. 

\begin{figure}[H]
\centering
    \begin{tabular}{c}
    \includegraphics[width=0.9\textwidth]{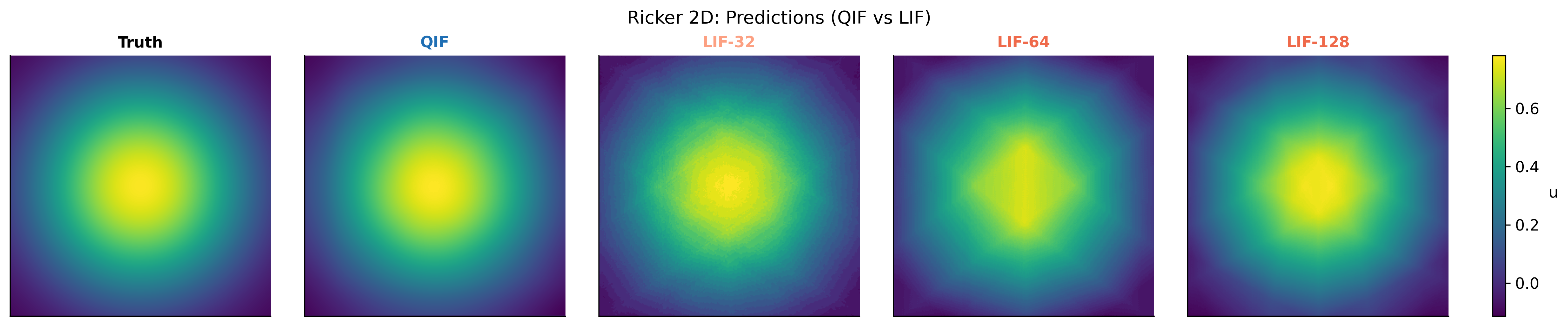} \\
    \includegraphics[width=0.9\textwidth]{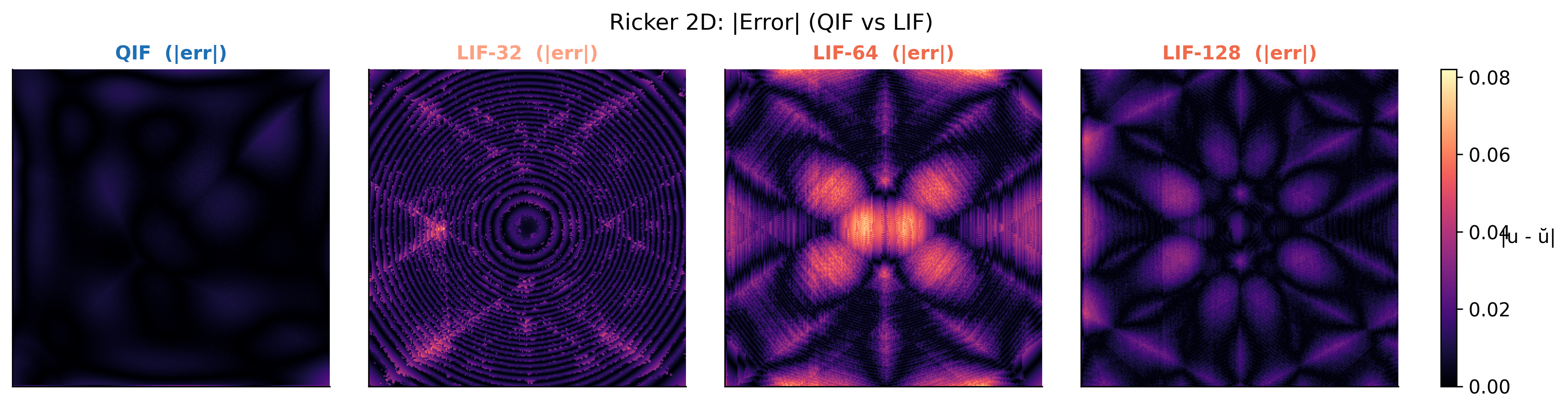} \\
   
     \begin{tabular}{cc}
      \includegraphics[width=0.45\textwidth]{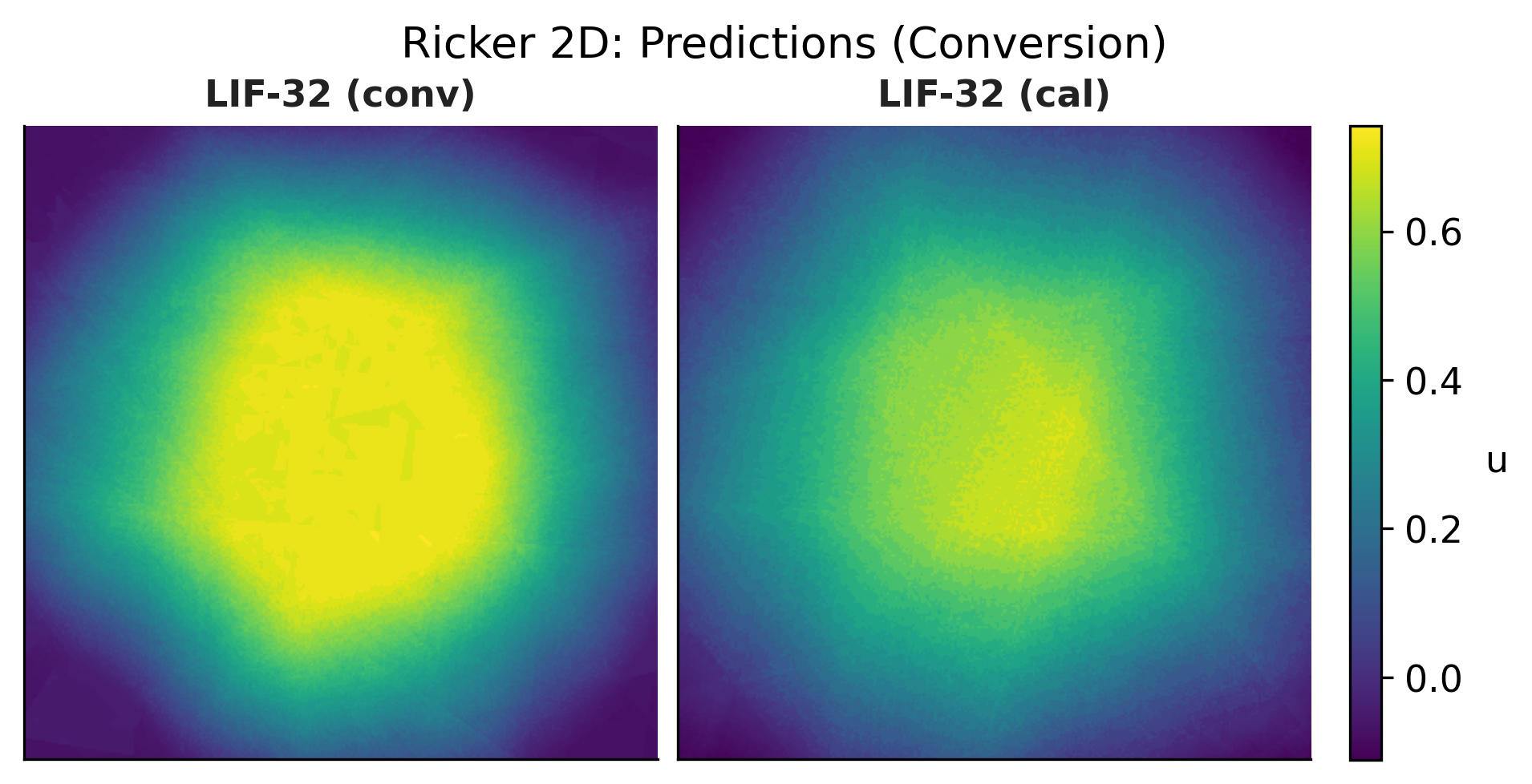} &
     \includegraphics[width=0.45\textwidth]{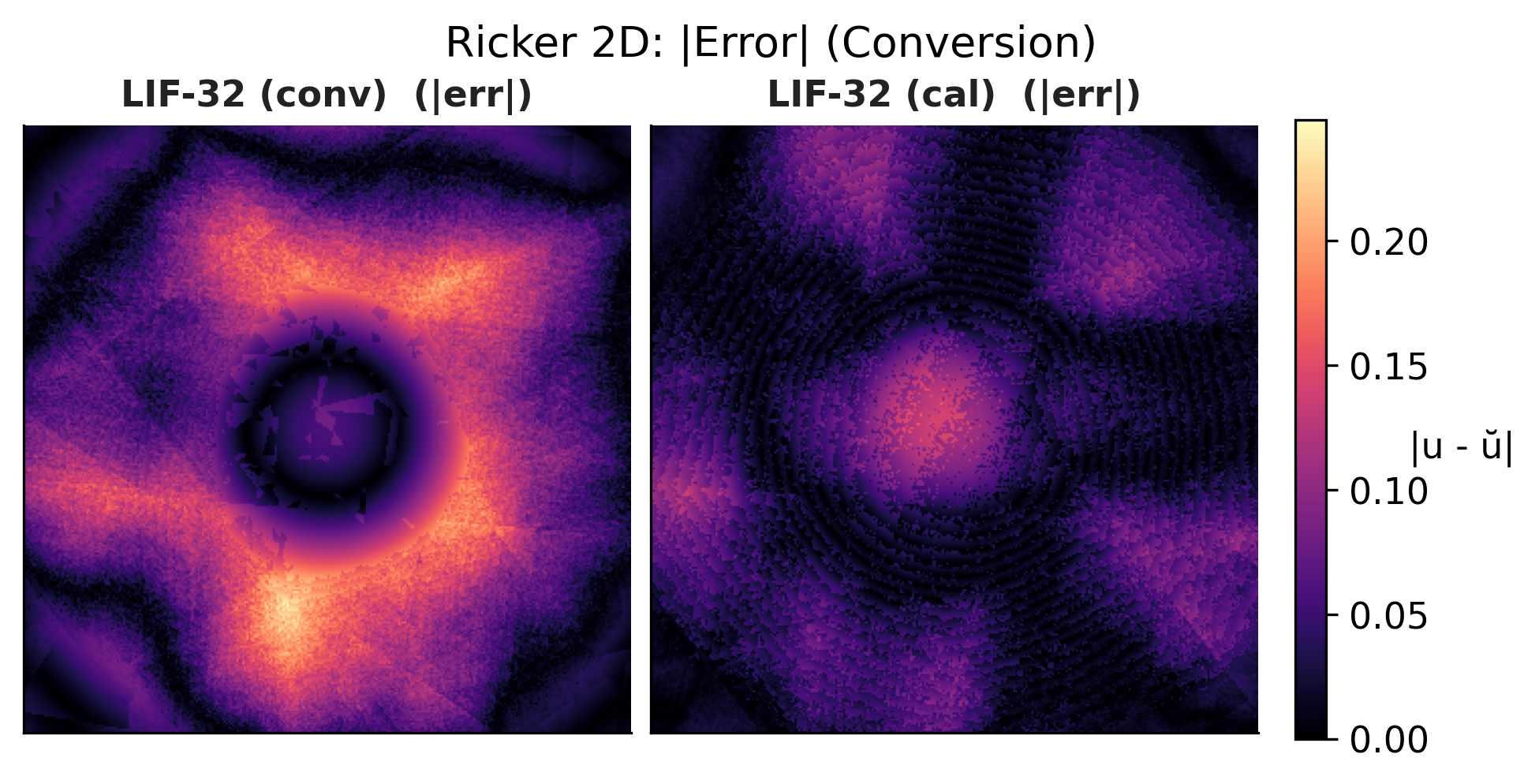} \\
    
     \end{tabular}

\end{tabular}
  \caption{Ricker wavelet results: Comparison of predictions from QIF training, LIF direct training and conversion. The first row is comparison of QIF with LIF direct training and the second row shows their errors. The third row is the prediction and errors of conversion with 32 simulation steps.}
\label{fig:ricker}
\end{figure}

The ground truth field, together with predictions from QIF with simulation length $T=2.0$ and LIF networks trained with 32, 64, and 128 simulation steps, are shown in the first row of Fig.~\ref{fig:ricker}. The corresponding absolute error maps are presented in the second row. It can be observed that the QIF neuron produces the smallest overall error and yields smooth, well-behaved predictions, in sharp contrast to the noisy and discontinuous outputs obtained from LIF neurons trained directly. The best performance among directly trained LIF models is achieved with 128 simulation steps. However, its error remains more than twice that of the QIF model, as shown in Table~\ref{tab:metrics}.

For ANN-to-SNN conversion, the performance is inconsistent: while conversion with 32 time steps provides moderately accurate predictions, increasing the number of simulation steps even degrades the results. Therefore, only the conversion predictions and their errors for the 32-step case are displayed in the third row of Fig.~\ref{fig:ricker}. As shown in Table~\ref{tab:metrics}, in this case the uncalibrated model yields a relative $L_2$ error of 25\%, which is only reduced to 12.99\% after calibration. Despite this improvement, the calibrated model still exhibits an error nearly ten times larger than that of the QIF model, which is the best accuracy achievable through conversion.

These results highlight the stability and smoothness advantages of QIF neurons in higher-dimensional regression tasks.

\subsubsection{Ripple-like function}

Another two-dimensional regression problem involving a ripple-like function defined as
\[
u(x, y) = 0.25 \sin\!\left(4\pi \sqrt{x^2 + y^2}\right) + 0.5,
\]
is also evaluated.
The quantitative results are summarized in Table~\ref{tab:metrics}. Here the simulation time for QIF is again $T=2.0$. Consistent with the previous examples, the QIF neuron demonstrates superior accuracy with relative $L_2$ error of 1.46\% compared to the LIF-based models whose best relative $L_2$ error is 2.43\%, confirming QIF's advantage in learning oscillatory functions.

\subsection{Deep neural operator}

\subsubsection{1D Poisson equation}
To evaluate the performance of QIF neurons in operator learning, we apply the spiking DeepONet architecture to a one-dimensional Poisson equation of the form
\[
- u_{xx} = g(x), \quad x\in[-1,1],\quad 
u(-1)=u(1)=0,
\]
where the source term $g(x)$ is sampled from a Gaussian random field (GRF) using the RBF kernel $k_\text{RBF}(x,x') = exp(-\frac{(x-x')^2}{2})$. The DeepONet consists of a branch network with architecture [51, 64, 64, 128] and a trunk network with architecture [1, 64, 64, 128]. A total of 800 samples are used for training and 800 for testing. Both inputs and outputs are normalized prior to training. The batch size is set to 50, and the model is trained for 1500 epochs. For the QIF neuron, the simulation time is $T = 3.0$. 

\begin{figure}[H]
\centering
    \begin{tabular}{ccc}
  \includegraphics[width=0.33\textwidth]{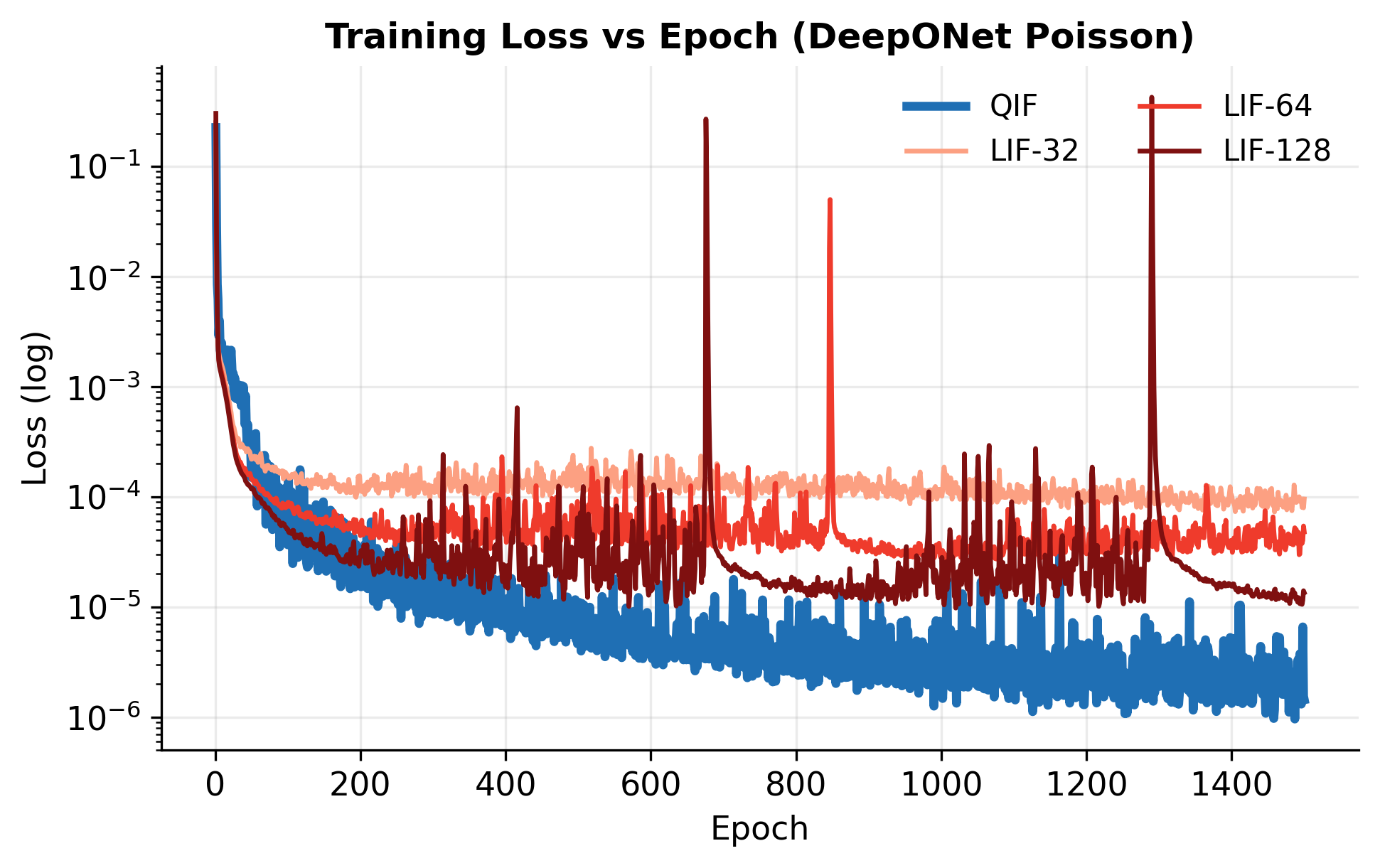} &
  \includegraphics[width=0.34\textwidth]{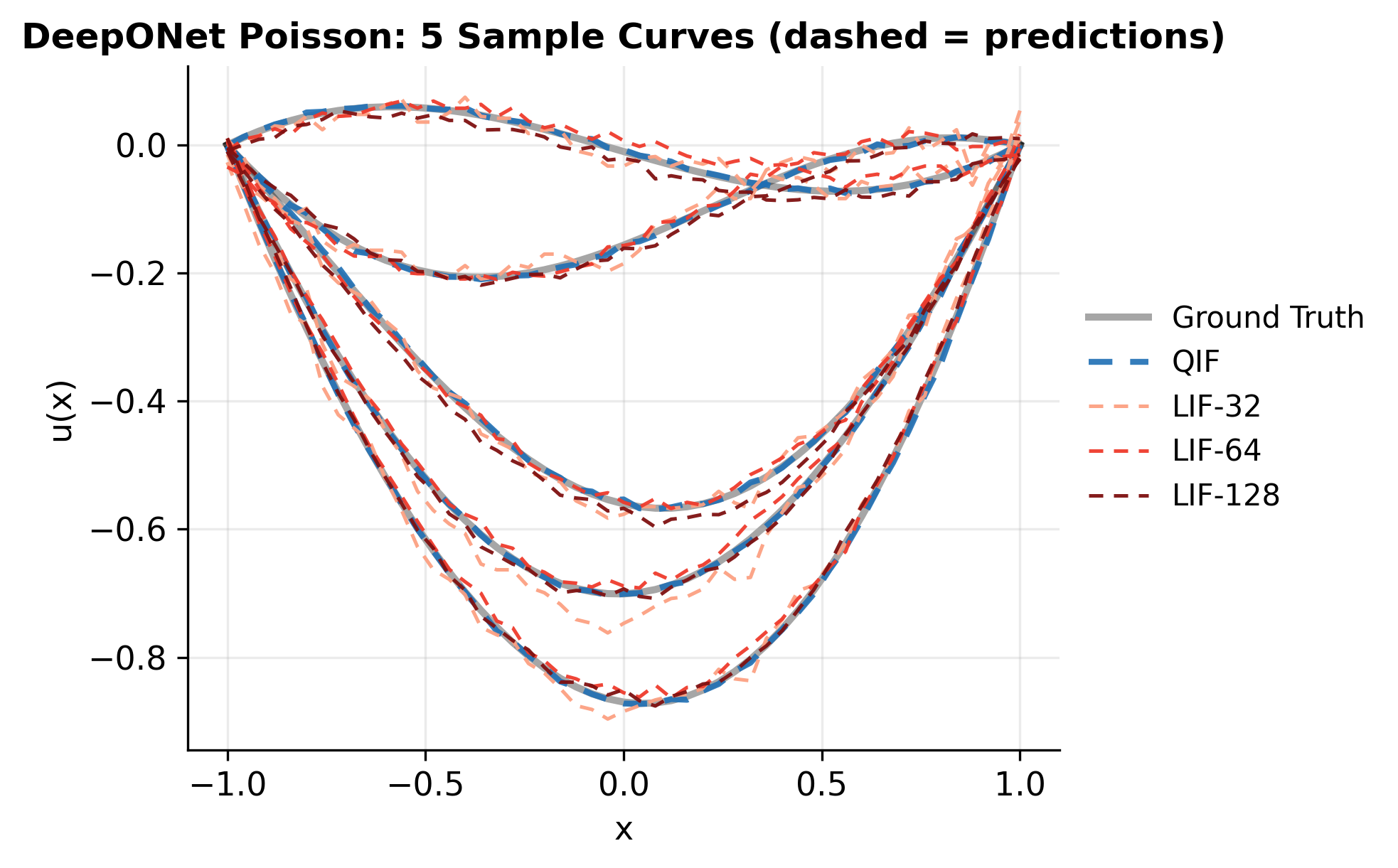} &
   \includegraphics[width=0.32\textwidth]{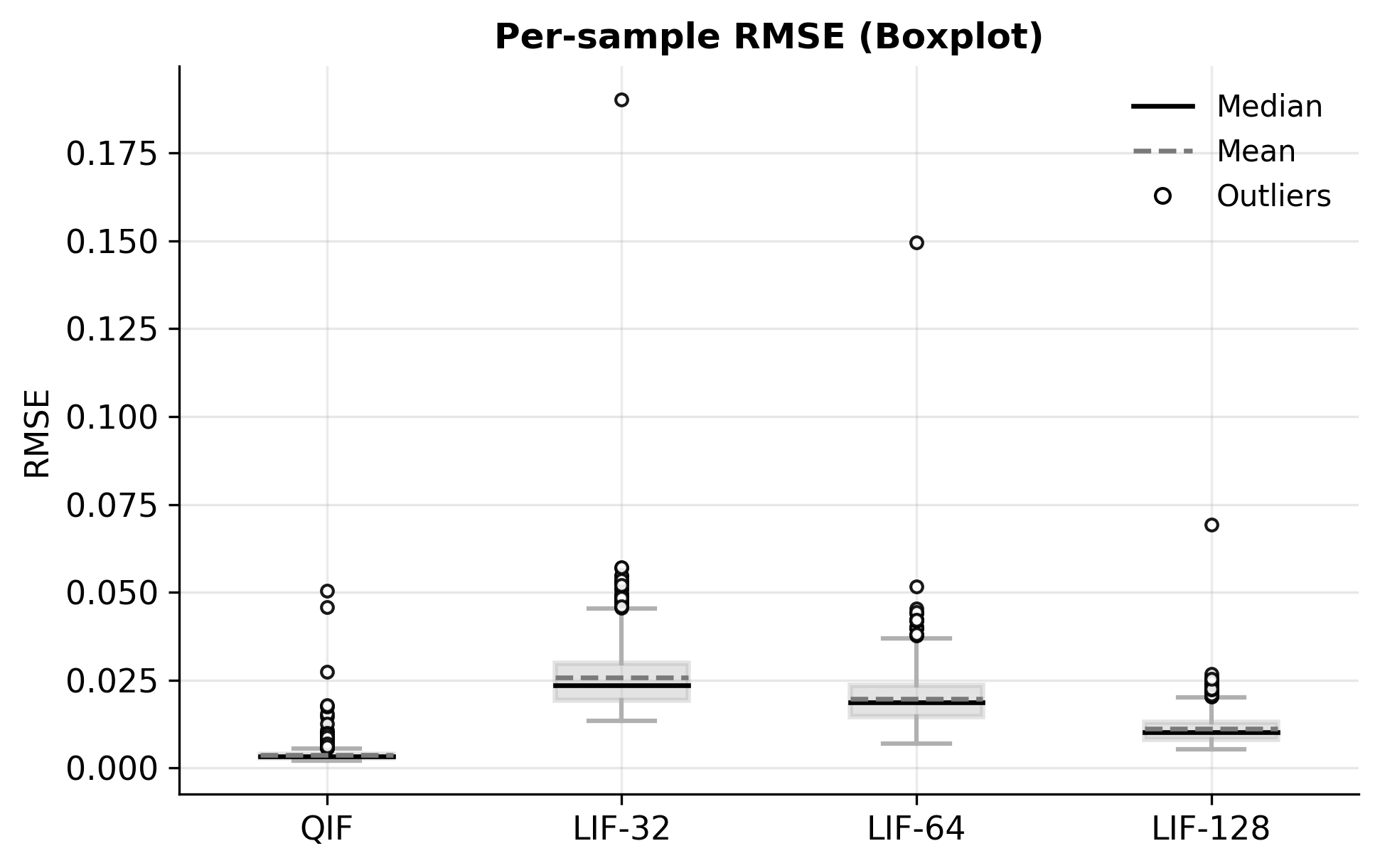}\\
   (a) Loss history & (b) Predictions & (c) RMSE \\
\end{tabular}
  \caption{DeepONet results: (a): Loss history in training with QIF neuron, LIF neuron with 32 simulation steps (LIF-32), LIF neuron with 64 simulation steps (LIF-64), LIF neuron with 128 simulation steps (LIF-128). (b): Parabola ground truth and predictions using QIF, LIF-32, LIF-64 and LIF-128 neurons. (c): Box and whisker plot for RMSE across samples.}
\label{fig:deeponet}
\end{figure}

As shown in Fig.~\ref{fig:deeponet}, the QIF-based DeepONet achieves the most accurate and smooth predictions without the jagged artifacts observed in LIF-based networks. Moreover, the QIF model exhibits a smaller variance in RMSE across samples, demonstrating improved stability and generalization in the operator learning framework. As shown in Table \ref{tab:metrics}, the best performance LIF model and the QIF model achieve relative $L_2$ error 8.88\% and 2.73\%, respectively.

\subsection{Physics-informed neural networks}

Finally, the performance of QIF neurons in solving physics-constrained problems is tested. We apply PINN to a one-dimensional PDE, spatio-temporal (1+1)D as wells as (2+1)D PDEs.

\subsubsection{1D Poisson equation}

We first consider the one-dimensional Poisson equation,
\[
- u_{xx} = 2\pi^2 \sin(\pi x), \quad x\in[0,1], \quad u(0) = u(1) = 0,
\]
whose analytical solution is given by $u(x) = 2\sin(\pi x)$. The network architecture used 2 hidden layers with 64 neurons per layer, with 1000 collocation points for training and a batch size of 50. Training is performed for 500 epochs. To enforce boundary conditions, we apply a hard constraint by multiplying the network output by $x(1 - x)$. 

\begin{figure}[H]
\centering
    \begin{tabular}{cc}
  \includegraphics[width=0.49\textwidth]{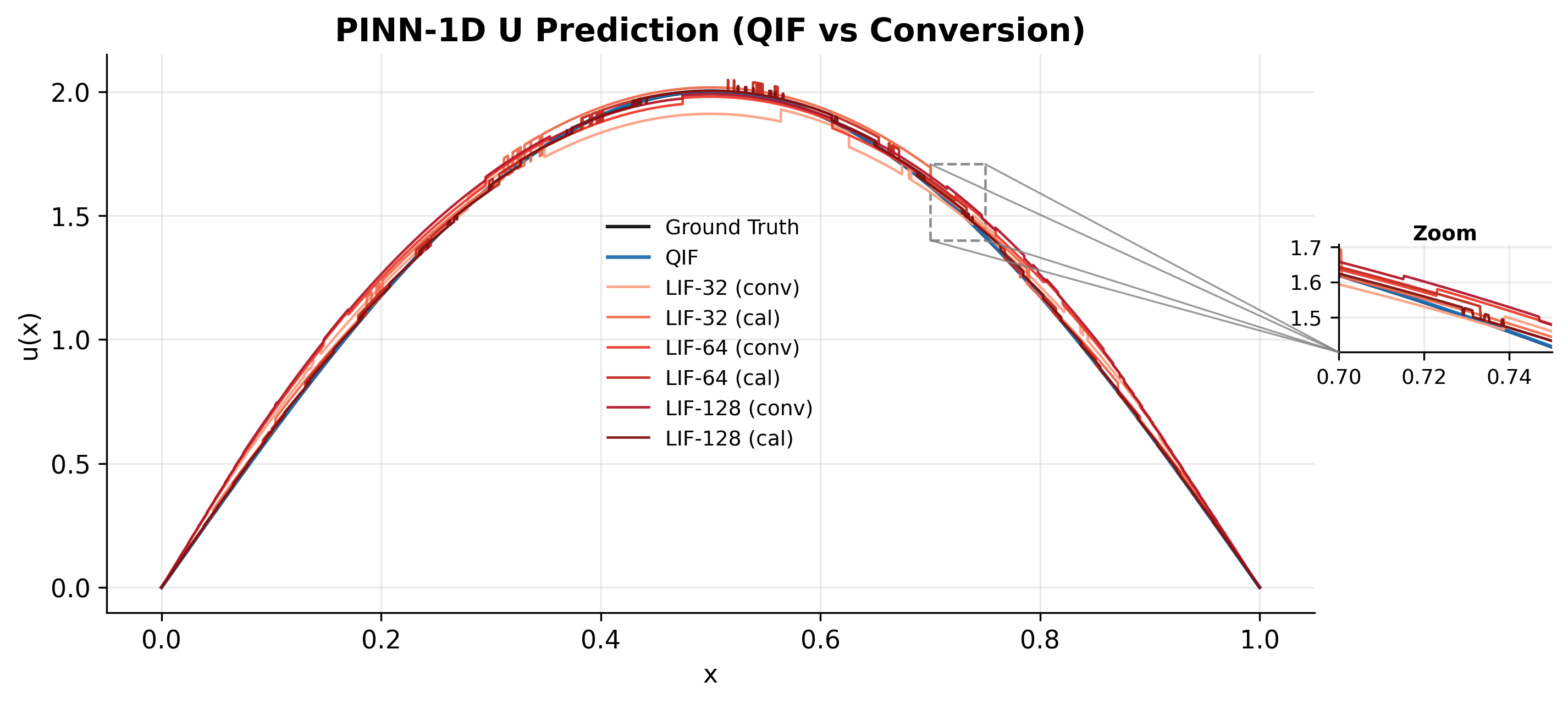} &
  \includegraphics[width=0.49\textwidth]{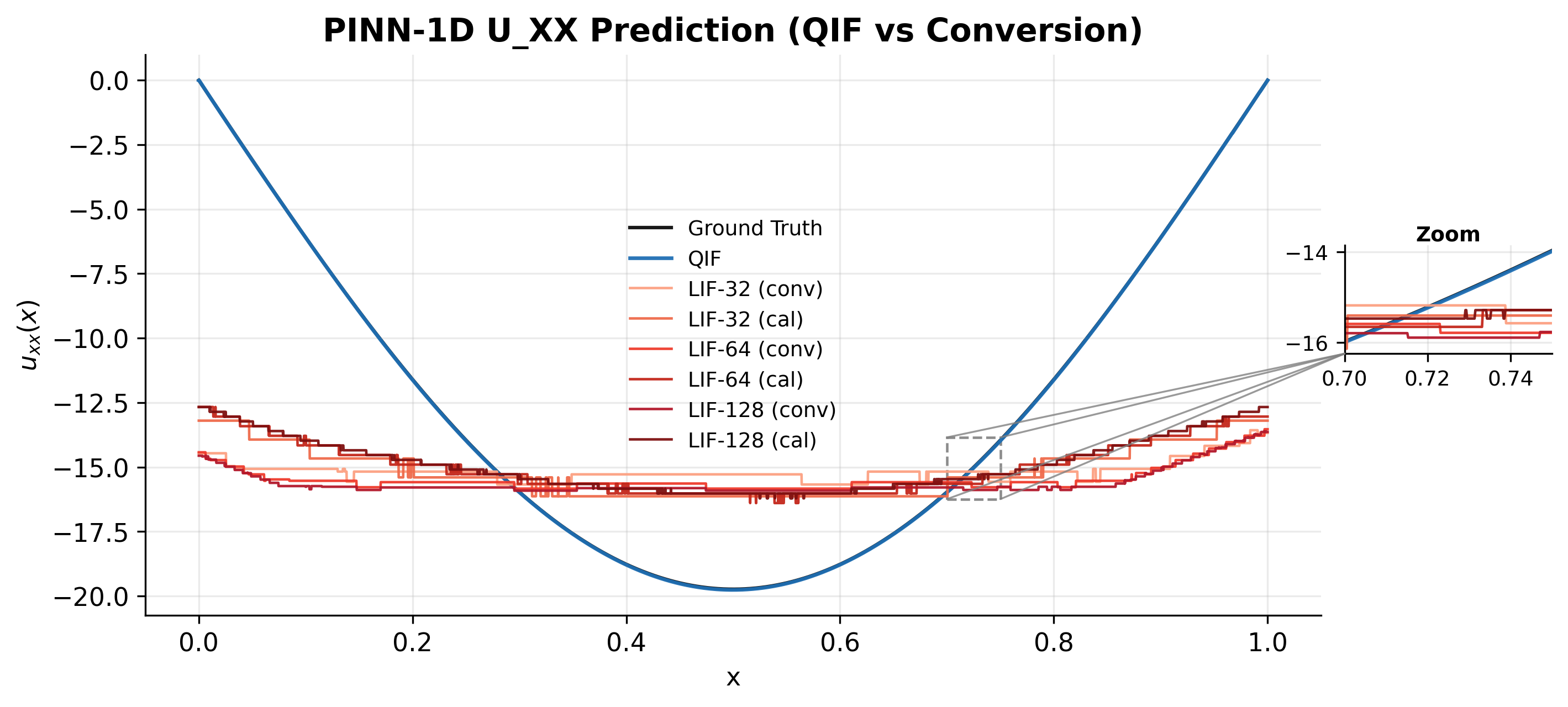}\\
    (a) Prediction of the solution. & (b) Prediction of the second derivative. \\
\end{tabular}
  \caption{PINN 1D Poisson results: Prediction from SNN using QIF traing and LIF conversion without calibration (conv) and with calibration (cali). (a): Prediction of the 1D Poisson solution. For LIF conversion, LIF-32, LIF-64 and LIF-128 correspond to 32, 64 and 128 simulation time steps respectively. (b): Prediction of the second derivative of $u$. Correct computation of $u_{xx}$ is critical in the success of learning a PINN by the PDE residual.}
\label{fig:pinn1d}
\end{figure}

Figure~\ref{fig:pinn1d}(a) compares the solutions obtained using direct QIF-based training with $T=2.0$ and ANN-to-SNN conversion with LIF neurons. For the converted models, we test 32, 64, and 128 simulation steps, both with and without calibration, denoted as “conv” for direct conversion and “cal” for calibrated conversion. Although calibration improves accuracy slightly, the resulting solutions remain non-smooth. In contrast, the QIF neuron produces a smooth and accurate prediction, whose error is only 0.13\% and closely follows the analytical solution.

Figure~\ref{fig:pinn1d}(b) shows the corresponding second derivative $u_{xx}$, which is central to computing the PDE residual loss in PINN. The QIF-based model accurately captures this second derivative, while all LIF-based converted SNNs deviate significantly from the true profile. This result indicates that, although ANN-to-SNN converted models can approximate the solution, they fail to preserve the physics constraints inherent in PINNs. The QIF neuron, on the other hand, successfully maintains both accuracy and physical consistency.

\subsubsection{(1+1)D Burgers equation}

Next, the spiking PINN framework is performed on the (1+1)D viscous Burgers equation, defined as
\[
u_t + u u_x - \nu u_{xx} = 0, \quad x \in [-1, 1], \quad t \in [0, 1],
\]
where the viscosity coefficient is set to $\nu = \frac{0.01}{\pi}$. The initial condition is $u(0, x) = -\sin(\pi x)$ and the boundary conditions are $u(t, -1) = u(t, 1) = 0$. The neural network architecture used is $6\times 64$. A total of 6579 collocation points are employed for training. The model is trained for 15{,}000 epochs, during which the QIF-based PINN with simulation time $T=2.0$ achieves a relative $L_2$ error of 2.42\%.

\begin{figure}[H]
\centering
    \begin{tabular}{c}
    \includegraphics[width=0.55\textwidth]{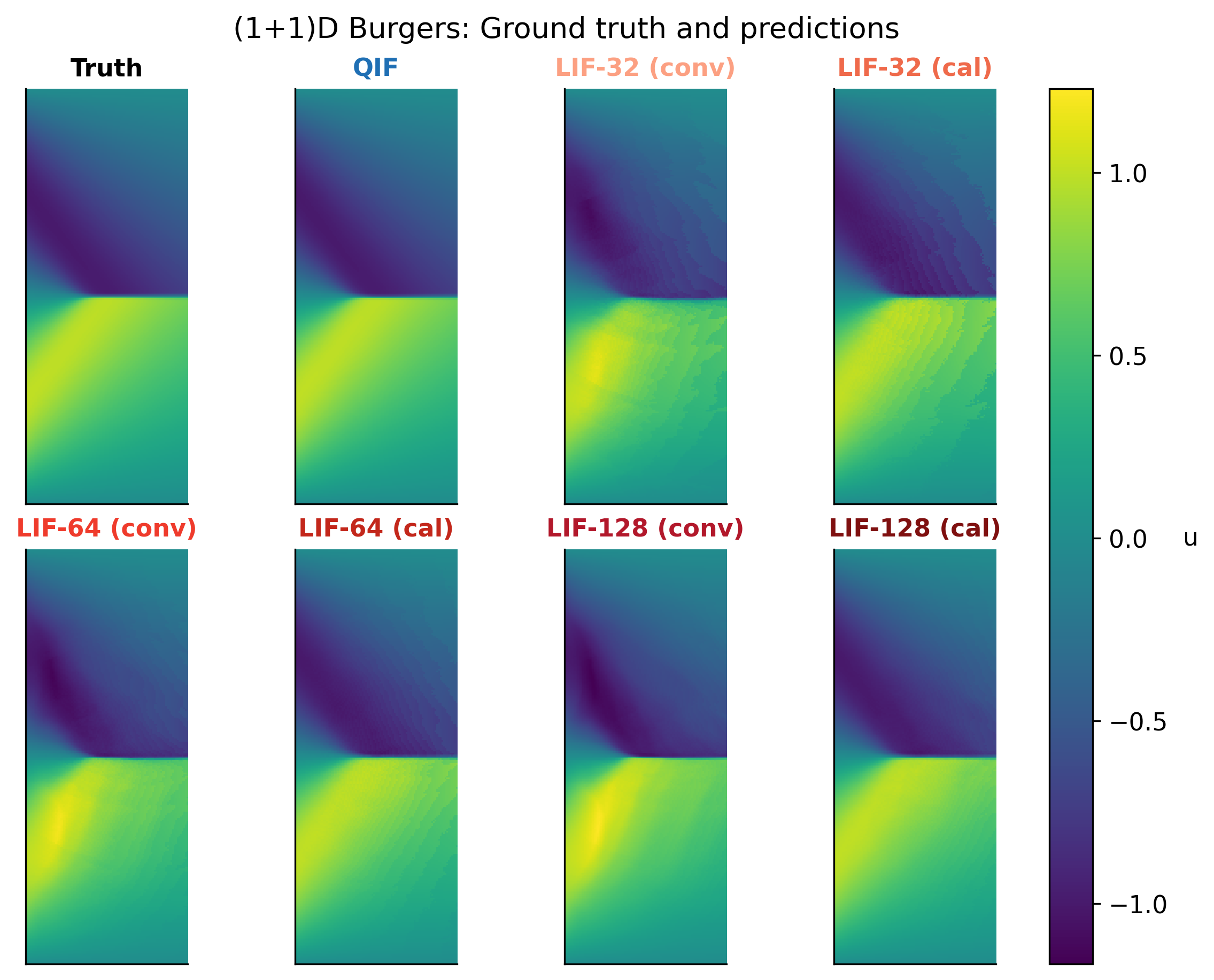} \\
     (a) \\
  \includegraphics[width=0.55\textwidth]{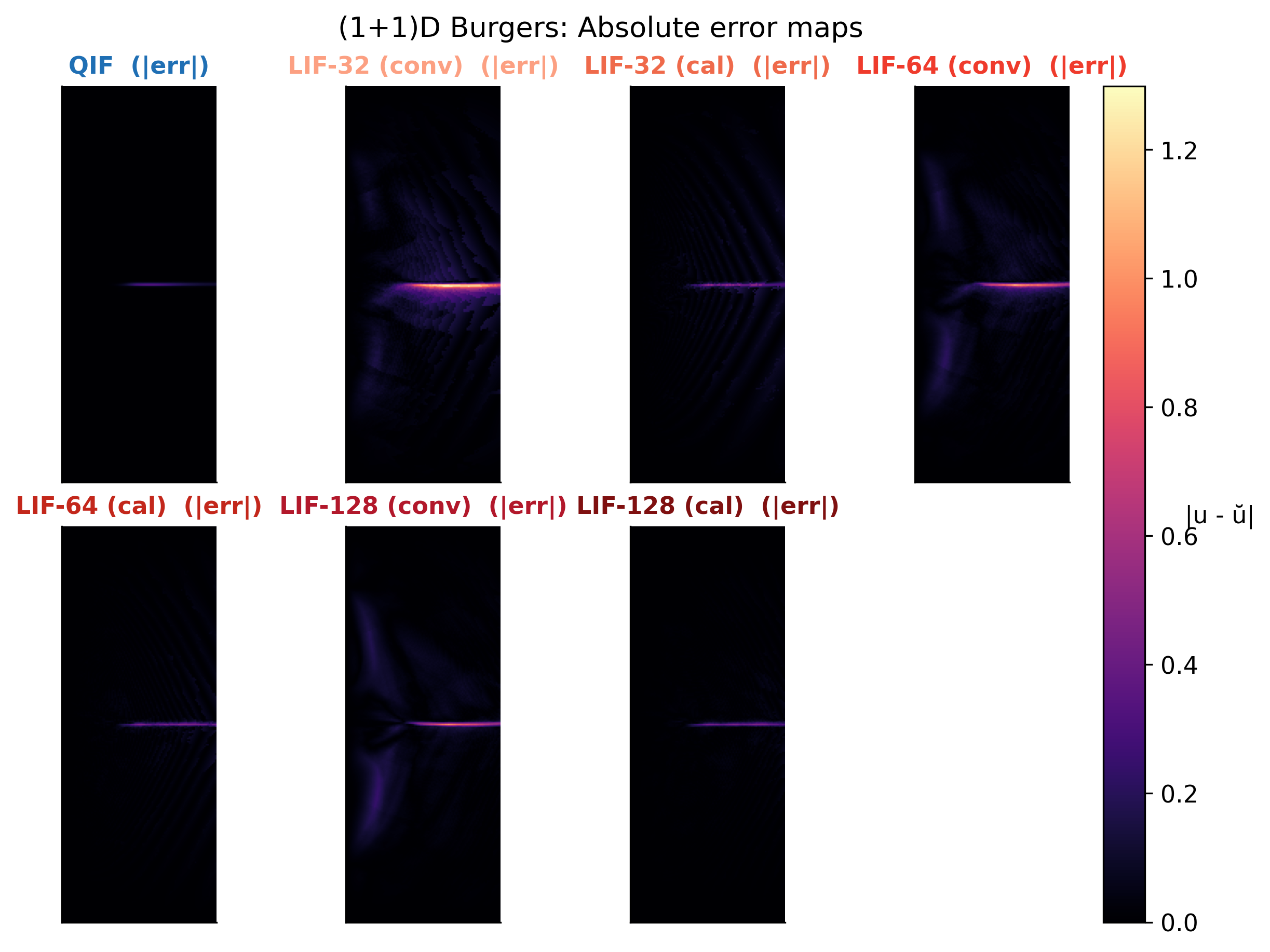} \\
   (b)\\ 
    \includegraphics[width=0.55\textwidth]{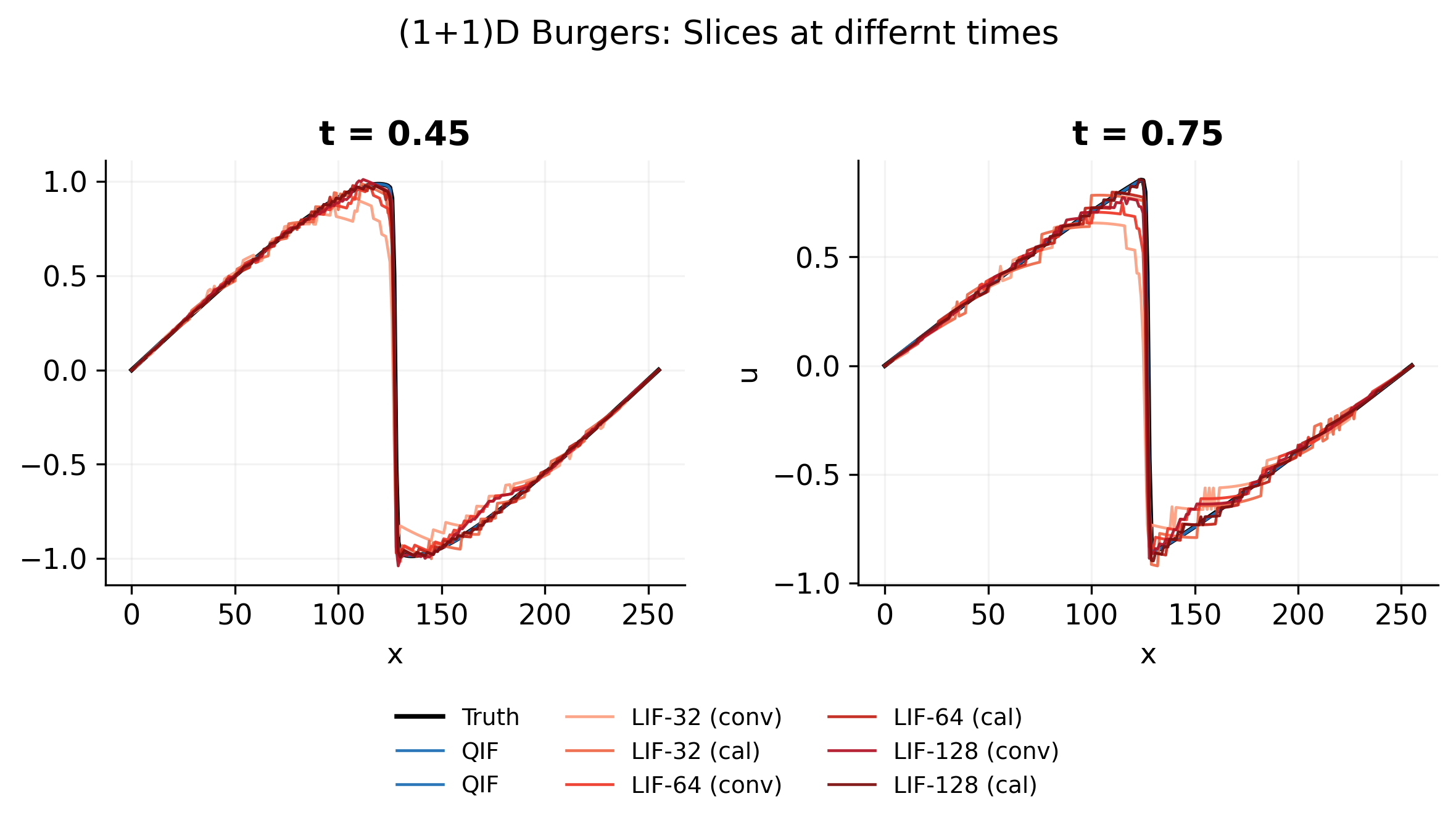}\\
     (c)
\end{tabular}
  \caption{(1+1)D Burgers results.}
  
\label{fig:pinn_burger}
\end{figure}

As illustrated in Fig.~\ref{fig:pinn_burger}(a), increasing the number of simulation steps in LIF-based models leads to smoother predictions, and calibration further improves the results. However, Fig.~\ref{fig:pinn_burger}(c) shows that even with 128 simulation steps and calibration, the converted LIF models still produce jagged solution curves. Moreover, they fail to accurately capture the sharp gradient characteristic of the viscous Burgers solution. In contrast, the QIF neuron provides a smooth and physically consistent prediction that closely follows the analytical profile. Figure~\ref{fig:pinn_burger}(b) further confirms that the QIF model yields the smallest point-wise errors across the entire spatiotemporal domain.

\subsubsection{(2+1)D Navier-Stokes equation} 
To examine the QIF PINN performance further, we test with the (2+1)D Beltrami flow. The governing incompressible Navier-Stokes equations are:
$$u_{t} + (u \cdot u_x + v \cdot u_y) = -p_x + (u_{xx}+u_{yy}),$$
$$v_{t} + (u \cdot v_x + v \cdot v_y) = -p_y + (v_{xx}+v_{yy}),$$
$$u_x + v_y = 0$$
for $x \in [-1,1], y \in [-1,1], t\in[0,1]$. The initial and boundary conditions are
$$u(x,y,0) = -cos(x)sin(y), v(x,y,0) = sin(x)cos(y),$$ 
$$u(\pm 1, y, t) = -cos(1)sin(y)e^{-2t}, u(x, \pm 1, t) = \mp cos(x)sin(1)e^{-2t},$$
$$v(\pm 1, y, t) = \pm cos(1)cos(y)e^{-2t}, v(x, \pm 1, t) = sin(x)cos(1)e^{-2t}.$$

\begin{figure}[H]
\centering
    \begin{tabular}{c}
    \includegraphics[width=0.8\textwidth]{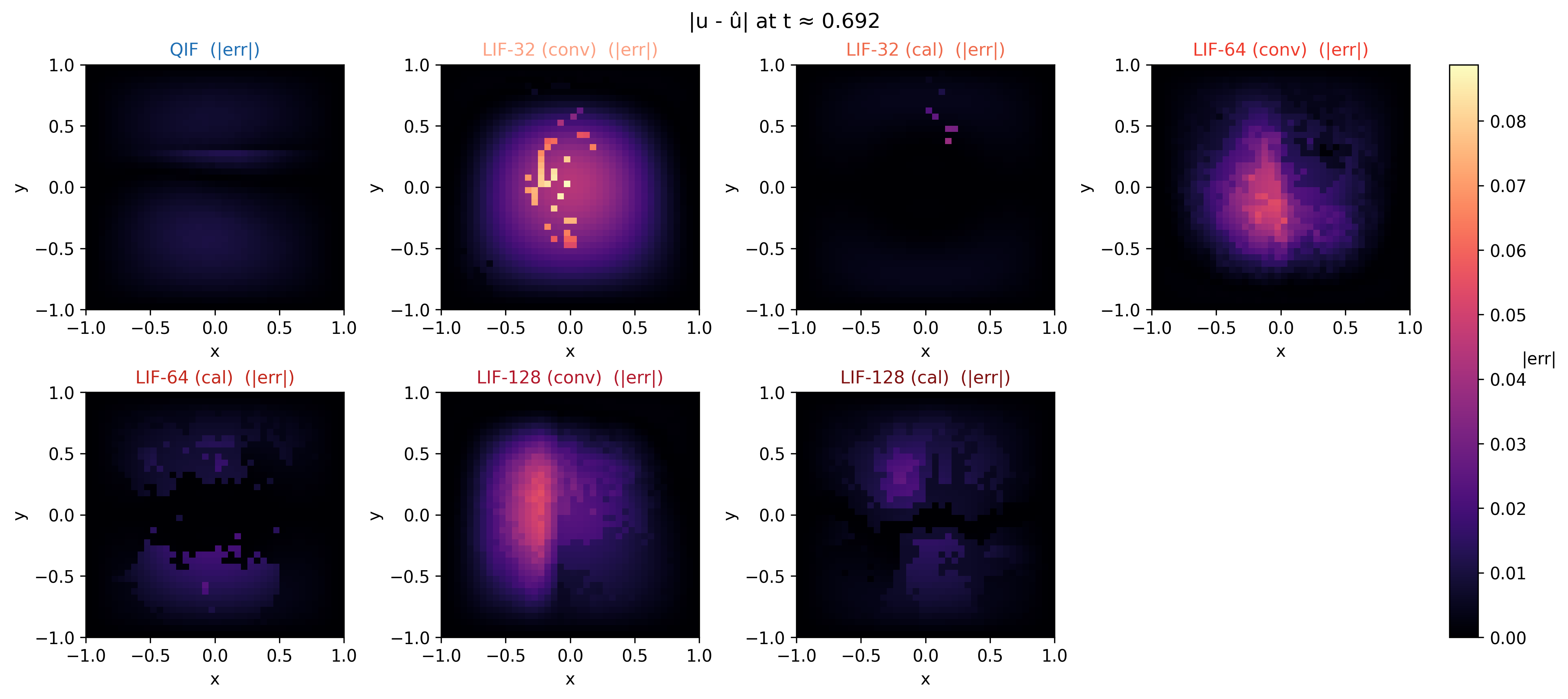} \\
     (a) \\
  \includegraphics[width=0.8\textwidth]{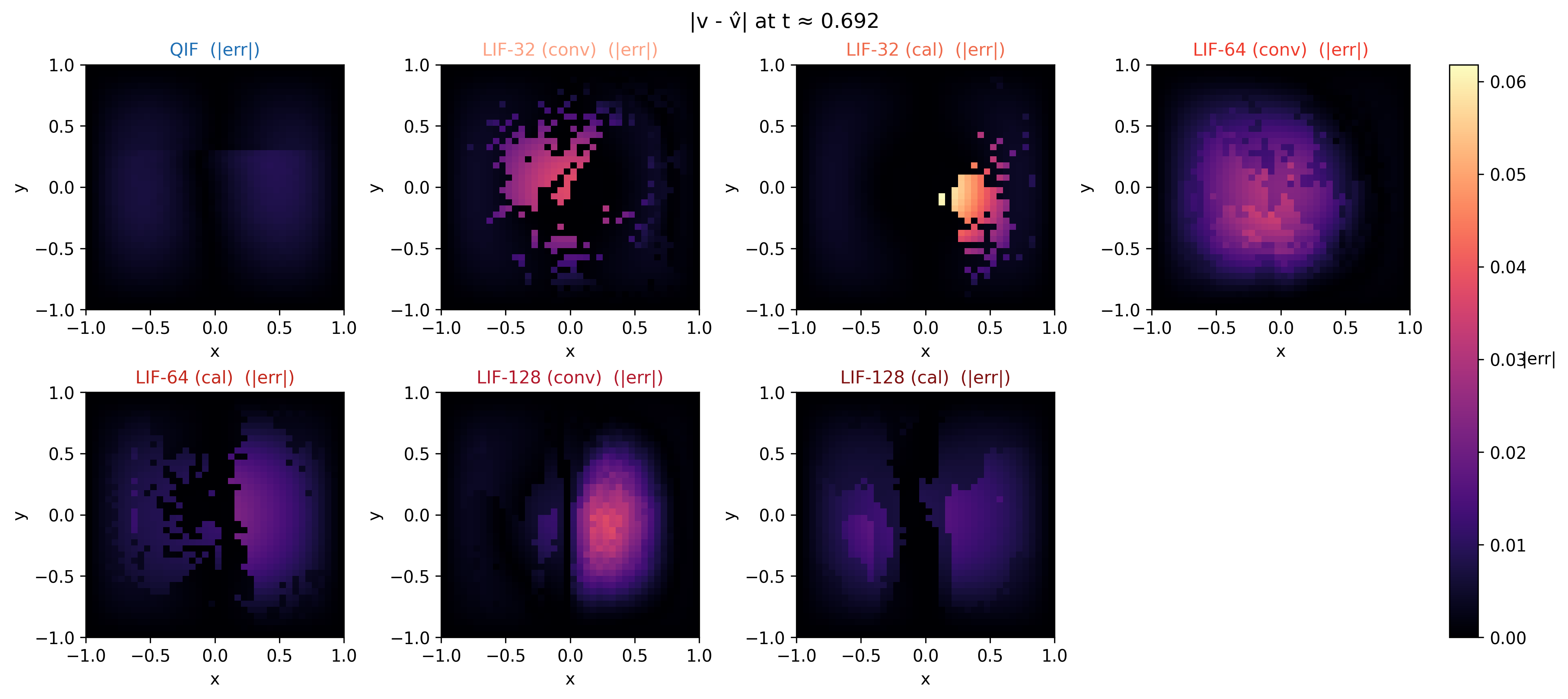} \\
   (b)\\ 
 
\end{tabular}
  \caption{(2+1)D Beltrami flow prediction error of (a) $u$ and (b) $v$ at one snapshot.}
\label{fig:pinn_ns}
\end{figure}

We use a PINN with size $6 \times 64$. The metrics in Table \ref{tab:metrics} still shows that QIF with simulation time of $2.0$ gives a relative $L_2$ error of 4.01\% for $u$ and 3.30\% for $v$. Although LIF conversion with 64 simulation time steps and conversion achieves a slightly lower  error of 3.55\% for $u$, the error for $v$ is much higher which is 4.49\%. For visualization, the prediction error plots of $u$ and $v$ at $t\approx 0.692$ for QIF and LIF are illustrated in in Figure \ref{fig:pinn_ns}. It could also be seen  from the error plot that the predictions from conversion are not smooth, even for LIF-64 with calibration which achieves a smaller error.

For a comprehensive summary of all quantitative results across experiments, please refer to Table~\ref{tab:metrics}.

\definecolor{lightblue}{RGB}{220, 235, 255}
\definecolor{lightgray}{gray}{0.9}

\begin{table}[h!]
  \centering
  \begin{threeparttable}
  \caption{Test error comparison across problems. We include estimates of the maximum absolute error (MAE), the root mean square error (RMSE) and the relative $L_2$ error.}
  \label{tab:metrics}
  \setlength{\tabcolsep}{4.5pt}
  \begin{tabular}{
    l S S S S  S S S S
  }
    \hline
    & \multicolumn{4}{c}{\textbf{Parabola fitting (1D)}} & \multicolumn{4}{c}{\textbf{Ricker wavelet fitting (2D)}}\\
    \cmidrule(lr){2-5} \cmidrule(lr){6-9}
    \textbf{Model} & {MAE} & {RMSE} & {Rel $L_2$(\%)} & {$R^2$}
                   & {MAE} & {RMSE} & {Rel $L_2$(\%)} & {$R^2$} \\
    \hline
    \rowcolor{lightblue}
    QIF                 & 0.001 & 0.003 & 0.609 & 0.999 & 0.004 & 0.005 & 1.31 & 0.999  \\
    LIF–32 (conv)       & 0.081 & 0.099 & 22.172 & 0.889 & 0.075 & 0.091 & 25.10 & 0.843 \\
    LIF–32 (cali)       & 0.030 & 0.034 & 7.683 & 0.987 & 0.037 & 0.047 & 12.99 & 0.958 \\
    LIF–32 (direct)     & 0.011 & 0.014 & 3.118 & 0.998 & 0.012 & 0.015 & 4.09 & 0.996 \\
    LIF–64 (conv)       & 0.072 & 0.094 & 20.947 & 0.901 & N/A & N/A & N/A & N/A \\
    LIF–64 (cali)       & 0.018 & 0.022 & 4.984 & 0.994 & N/A & N/A & N/A & N/A \\
    LIF–64 (direct)     & 0.006 & 0.008 & 1.803 & 0.999 & 0.018 & 0.024 & 6.51 & 0.989 \\
    LIF–128 (conv)       & 0.085 & 0.102 & 22.743 & 0.884 & N/A & N/A & N/A & N/A \\
    LIF–128 (cali)       & 0.015 & 0.021 & 4.582 & 0.995 & N/A & N/A & N/A & N/A \\
    LIF–128 (direct)    & 0.005 & 0.007 & 1.451 & 0.999 & 0.009 & 0.012 & 3.29 & 0.997 \\

    \hline
    & \multicolumn{4}{c}{\textbf{Ripple fitting (2D)}} & \multicolumn{4}{c}{\textbf{Poisson solution operator learning (1D)}}\\
    \cmidrule(lr){2-5} \cmidrule(lr){6-9}
    \textbf{Model} & {MAE} & {RMSE} & {Rel $L_2$(\%)} & {$R^2$}
                   & {MAE} & {RMSE} & {Rel $L_2$(\%)} & {$R^2$} \\
    \hline
    \rowcolor{lightblue}
    QIF                 & 0.006 & 0.008 & 1.46 & 0.998 & 0.003 & 0.004 & 2.73 & 0.995  \\
    LIF–32 (cal)       & 0.054 & 0.066 & 12.64 & 0.857 & N/A & N/A & N/A & N/A \\
    LIF–64 (cal)       & 0.050 & 0.063 & 12.01 & 0.871 & N/A & N/A & N/A & N/A \\
    LIF–128 (cal)       & 0.030 & 0.042 & 8.03 & 0.942 & N/A & N/A & N/A & N/A \\
    LIF–32 (direct)     & 0.011 & 0.013 & 2.73 & 0.993 & 0.021 & 0.026 & 19.77 & 0.721 \\
    LIF–64 (direct)     & 0.008 & 0.010 & 1.86 & 0.997 & 0.017 & 0.020 & 15.72 & 0.815 \\
    LIF–128 (direct)    & 0.010 & 0.013 & 2.43 & 0.995 & 0.010 & 0.011 & 8.88 & 0.942 \\
    \hline
    & \multicolumn{4}{c}{\textbf{Poisson solving (1D)}} & \multicolumn{4}{c}{\textbf{Burgers solving ((1+1)D)}}\\
    \cmidrule(lr){2-5} \cmidrule(lr){6-9}
    \textbf{Model} & {MAE} & {RMSE} & {Rel $L_2$(\%)} & {$R^2$}
                   & {MAE} & {RMSE} & {Rel $L_2$(\%)} & {$R^2$} \\
    \hline
    \rowcolor{lightblue}
    QIF                 & 0.002 & 0.002 & 0.13 & 0.999 & 0.002 & 0.015 & 2.42 & 0.999  \\
    LIF–32 (conv)       & 0.045 & 0.052 & 3.64 & 0.993 & 0.047 & 0.106 & 17.19 & 0.970 \\
    LIF–32 (cal)     & 0.025 & 0.031 & 2.21 & 0.997 & 0.021 & 0.039 & 6.38 & 0.996 \\
    LIF–64 (conv)     & 0.042 & 0.050 & 3.55 & 0.993 & 0.034 & 0.072 & 11.64 & 0.986 \\
    LIF–64 (cal)     & 0.013 & 0.017 & 1.18 & 0.999 & 0.012 & 0.030 & 4.83 & 0.998 \\
    LIF–128 (conv)    & 0.046 & 0.057 & 4.053 & 0.991 & 0.030 & 0.062 & 10.15 & 0.990 \\
    LIF–128 (cal)    & 0.007 & 0.010 & 0.679 & 0.999 & 0.009 & 0.026 & 4.28 & 0.998 \\
    \hline
    & \multicolumn{4}{c}{\textbf{Navier-Stokes solving ((2+1)D) $u$}} & \multicolumn{4}{c}{\textbf{Navier-Stokes solving ((2+1)D) $v$}}\\
    \cmidrule(lr){2-5} \cmidrule(lr){6-9}
    \textbf{Model} & {MAE} & {RMSE} & {Rel $L_2$(\%)} & {$R^2$}
                   & {MAE} & {RMSE} & {Rel $L_2$(\%)} & {$R^2$} \\
    \hline
    \rowcolor{lightblue}
    QIF                 & 0.006 & 0.009 & 4.01 & 0.998 & 0.005 & 0.007 & 3.30 &  0.999 \\
    LIF–32 (conv)       & 0.013 & 0.019 & 8.50 & 0.993 & 0.006 & 0.011 & 4.87 & 0.998 \\
    LIF–32 (cal)     & 0.005 & 0.009 & 4.03 & 0.998 & 0.006 & 0.011 & 4.92 & 0.998 \\
    LIF–64 (conv)     & 0.009 & 0.014 & 6.40 & 0.996 & 0.008 & 0.011 & 4.93 & 0.998 \\
    LIF–64 (cal)     & 0.005 & 0.008 & 3.55 & 0.999 & 0.007 & 0.010 & 4.49 & 0.998 \\
    LIF–128 (conv)    & 0.011 & 0.016 & 7.29 & 0.995 & 0.008 & 0.012 & 5.43 & 0.997 \\
    LIF–128 (cal)    & 0.007 & 0.010 & 4.22 & 0.998 & 0.007 & 0.010 & 4.35 & 0.998 \\
    \hline
    
  \end{tabular}

  \end{threeparttable}
\end{table}




\section{Summary and Future Directions}

In this work, we integrated the quadratic integrate-and-fire (QIF) neuron into regression-based scientific machine learning frameworks, demonstrating its advantages in achieving smooth and differentiable dynamics. Comprehensive comparisons were conducted across multiple benchmarks and architectures, including MLPs, DeepONets and PINNs, against state-of-the-art regression approaches using leaky integrate-and-fire (LIF) neurons. The results consistently show that QIF neurons outperform LIF neurons in both accuracy and smoothness of predictions.

This study represents a foundational step toward incorporating QIF neurons into broader scientific machine learning applications. The demonstrated smooth gradient behavior of QIF neurons highlights their potential for integration into more complex and large-scale spiking architectures. Future work will focus on extending QIF-based models to neuromorphic hardware implementations and exploring their capabilities in large-scale, event-driven computation for physics-informed and operator-learning tasks.


\section{Acknowledgements}
This work was supported by the Vannevar Bush Faculty Fellowship award (GEK) from ONR (N00014-22-1-2795). The work of PS and GEK is partially supported by the U.S. Department of Energy, Office of Science, Advanced Scientific Computing Research program under the Scalable, Efficient and Accelerated Causal Reasoning Operators, Graphs and Spikes for Earth and Embedded Systems (SEA-CROGS) project (Project No. 80278). Pacific Northwest National Laboratory is a multi-program national laboratory operated for the U.S. Department of Energy by Battelle Memorial Institute under Contract No. DE-AC05-76RL01830. 

{
\bibliographystyle{plain}
\bibliography{references}
}

\end{document}